\documentclass[runningheads]{llncs}

\usepackage[mobile]{eccv}

\usepackage{eccvabbrv}

\usepackage{graphicx}
\usepackage{booktabs}
\usepackage{array}
\usepackage{ragged2e}
\usepackage{multirow}
\usepackage{subcaption}
\usepackage{amsmath,amssymb}
\usepackage{algorithm}
\usepackage{algpseudocode}
\usepackage{titletoc}
\usepackage{CJKutf8}

\usepackage[accsupp]{axessibility}  

\usepackage{xcolor}
\usepackage[colorlinks=true,linkcolor=black,citecolor=eccvblue]{hyperref}
\usepackage{orcidlink}
\usepackage{wrapfig}
\usepackage{makecell}

\providecommand{\authcount}[1]{}

\begin{document}

\title{SignX: Continuous Sign Recognition in Compact Pose-Rich Latent Space}

\titlerunning{SignX: Sign Language Understanding in Latent Space}

\title{SignX: Continuous Sign Recognition in Compact Pose-Rich Latent Space}

\titlerunning{SignX: Sign Language Understanding in Latent Space}

\author{
Sen Fang\inst{1}\orcidlink{0009-0007-9463-4491} \and
Yalin Feng\inst{2}\orcidlink{0009-0000-8932-1545}\thanks{Equal Contribution} \and
Chunyu Sui\inst{3}\orcidlink{0009-0008-0497-3463} \and
Hongbin Zhong\inst{4}\orcidlink{0009-0003-2564-9674} \and
Yanxin Zhang\inst{5}\orcidlink{0009-0001-2307-901X} \and
Hongwei Yi\inst{6}\orcidlink{0000-0001-8669-2009} \and
Hezhen Hu\inst{7}\orcidlink{0000-0003-0327-1562} \and
Dimitris N. Metaxas\inst{1}\orcidlink{0000-0001-7142-7640}
}

\authorrunning{S.~Fang et al.}

\institute{
$^{1}$Rutgers University \quad
$^{2}$Nanyang Technological University \quad
$^{3}$Columbia University \quad
$^{4}$Georgia Institute of Technology \quad
$^{5}$University of Wisconsin--Madison \quad
$^{6}$Max Planck Institute for Intelligent Systems \quad
$^{7}$University of Texas at Austin \\
\url{https://SignerX.github.io/SignX}
\vspace{-24pt}
}
\begin{CJK*}{UTF8}{gbsn}
\maketitle

\begin{abstract}
The complexity of Sign Language (SL) data processing brings many challenges. The current approach to recognition of SL signs aims to translate RGB sign language videos through  pose information into Word-based ID Glosses, which serve to uniquely identify signs\footnote{Note that there is no shared convention for assigning such glosses to SL signs, so consistent glossing conventions must be used across all datasets.}. This paper proposes \textbf{SignX}, a novel framework for continuous sign language recognition (SLR) in compact pose-rich latent space. First, we construct a unified latent representation that encodes heterogeneous pose formats (SMPLer-X, DWPose, Mediapipe, PrimeDepth, and Sapiens Segmentation) into a compact, information-dense space. Second, we train a ViT-based Video-to-Pose module to extract this latent representation directly from raw videos. Finally, we develop a temporal modeling and sequence refinement method that operates entirely in this latent space. 
This multi-stage design achieves end-to-end SLR while significantly reducing computational consumption.
Experimental results demonstrate that SignX achieves SOTA accuracy on continuous SLR and Translation task, delivering nearly a 50-fold acceleration over pixel-space baselines.

\end{abstract}
\section{Introduction}
\label{sec:intro}

Sign Language Recognition (SLR) aims to automatically convert sign language videos into text or glosses (represented as upper-case text, which serve as unique identifiers of signs). It has important social value in facilitating barrier-free communication between deaf and hearing individuals \cite{camgoz2018neural, camgoz2020sign, stoll2020text2sign, yin-etal-2021-including-signed-languages}.
However, SLR faces technical challenges: \textcolor{purple}{(1)} \textit{Signed languages are complex multimodal movements}, involving hand and arm movements as well as facial expressions and body postures, with complex temporal dependencies among these modalities \cite{Bohacek_2022_WACV,signor,SIGNUM}. \textcolor{purple}{(2)}~\textit{Existing SL datasets are limited in size and lack a uniform preprocessing format}, with complicated processing pipelines, which severely constrains the application of deep learning methods \cite{duarte2021how2sign,JAsigning,camgoz2020sign}. 
\textcolor{purple}{(3)} \textit{Different pose processing formats attend to different visual aspects, leading to distributional shifts that limit effective integration} of innovative advantages across model types \cite{fang2025signllmsignlanguageproduction,fang2025signdiffdiffusionmodelamerican}.

\begin{wrapfigure}{r}{0.45\textwidth}
  \centering
  \includegraphics[width=0.99\linewidth]{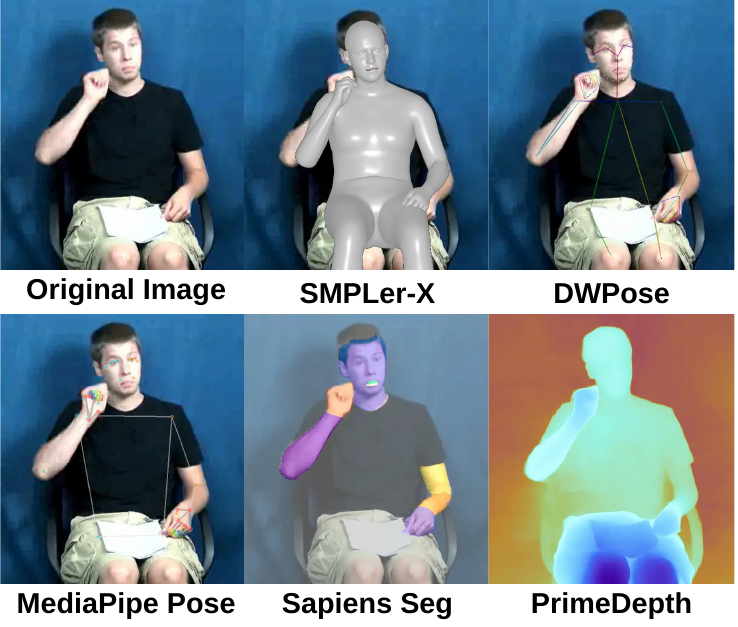}
  \vspace{-14pt}
  \caption{\textbf{Multimodal pose estimation methods:} SMPLer-X \cite{cai2023smplerx} can provide accurate 3D human body parameters; DWPose \cite{yang2023effective} focuses on real-time 2D keypoint detection; Mediapipe \cite{MediaPipe} provides lightweight but efficient 3D pose prediction; PrimeDepth \cite{zavadski2024primedepth} can obtain scene depth information; while Sapiens Segmentation \cite{khirodkar2024sapiens} provides fine-grained human body part segmentation results. These methods each have their own characteristics, providing rich feature representations for sign language recognition.}
  \vspace{-24pt}
  \label{fig:cover}
\end{wrapfigure}

To address these challenges, we develop a \textit{Vid2Pose module} by using a novel ViT model \cite{dosovitskiy2020vit}, which inspired by the five most powerful pose estimation models.
As shown in Fig. \ref{fig:cover}, we learn the Vid2Pose process in the encoder of the ViT model through a unified pose representation. Specifically, SMPLer-X \cite{cai2023smplerx} provides 3D body mesh with kinematic and pose parameters for modeling body dynamics;
DWPose \cite{yang2023effective} captures 2D keypoints with rich facial landmarks essential for grammatical expressions; MediaPipe \cite{MediaPipe} offers 3D joint coordinates that represent overall body motion trends. Additionally, PrimeDepth \cite{zavadski2024primedepth} provides spatial depth information for disambiguating front-back positioning, while Sapiens Segmentation \cite{khirodkar2024sapiens} captures human body shape through fine-grained part boundaries. This complementary design captures not only hand movements but also fine-grained details such as \textit{body dynamics, facial expressions, motion trajectories, spatial depth, and body shape}, with the goal of fully capturing a person's pose \cite{saunders2020progressive, stoll2020text2sign, pose_format_helper}.
This end-to-end processing approach greatly simplifies the \textit{pose representation} \cite{Bohacek_2022_WACV,cho2021unifying, chiang2019unified} process that is necessary for SL recognition.

Then we employ the \textit{Pose2Gloss method} to identify the pose features in the latent space, but most existing recognition models have several limitations: \textcolor{purple}{(1)} Most are based on \textit{raw pose data learning}, which is inconvenient in practical applications and requires assistance from other pose information extraction models \cite{wang2018video, wang2018high}. \textcolor{purple}{(2)}~Furthermore, models capable of real-time translation have \textit{low accuracy}, while highly accurate models have \textit{slow output speeds} \cite{saunders2021mixed}. \textcolor{purple}{(3)} Even purely pixel-space-based models are limited by \textit{high computational costs}, and their attention mechanisms cannot support higher performance ceilings for longer and more accurate translations \cite{cui2017recurrent, koller2020quantitative,gloss-informal,zelinka2020neural}.

To address these limitations, we develop a recognition framework that operates entirely in the compact pose-rich latent space. Starting from the 2048-dimensional pose features extracted by Vid2Pose, we employ a ResNet34 backbone \cite{7780459} followed by TemporalConv layers \cite{10.1007/978-3-319-49409-8_7} to capture hierarchical temporal patterns. These are jointly learned with the pose features.
%
To refine these predictions into coherent sequences, we apply a Transformer-based encoder-decoder \cite{zhang2023sltunet} that performs sequence-to-sequence refinement with beam search and CTC regularization \cite{camgoz2020sign, vaswani2017attention}. This design achieves accurate recognition directly in latent space, eliminating the need for raw pose processing and reducing computational overhead compared to vision-based approaches.

Through this multi-stage training, SignX achieves powerful functionality while maintaining architectural simplicity: It learns from various formats of prior knowledge and can directly use the raw video as input.
Experimental results show that SignX achieves good performance on our ASL and mainstream datasets \cite{asllrp2025signbank,forster2012rwth,Zhou2021ImprovingSLT-with-monolingual-CSLDaily,WLASL}, surpassing existing methods in both accuracy and robustness \cite{Chen-arxiv-2023-Robust}.

The main contributions of this paper can be summarized in the following three points:

\begin{itemize}
    \item We propose \textbf{SignX}, a novel framework for continuous sign language recognition that operates in a \textbf{compact pose-rich latent space}, unifying heterogeneous pose representations from five powerful estimation methods into a single information-dense encoding. 
    \item We develop a \textbf{ViT-based Vid2Pose module} that extracts unified pose representations (encompassing facial expressions, body dynamics, motion trajectories, spatial depth, and body shape) directly from raw videos in an end-to-end manner, eliminating the need for explicit pose estimation pipelines and significantly simplifying the workflow.
    \item We design a \textbf{latent-space recognition method} that combines ResNet temporal modeling with \textbf{Transformer-based sequence refinement}, achieving accurate sign recognition while reducing computational overhead compared to pixel-space approaches.
    
\end{itemize}
\section{Related Work}
\label{sec:related_short}

\textbf{Sign Recognition.}
While traditional SLR frameworks predominantly rely on feature extraction from high-dimensional pixel streams or raw skeletal coordinates, these approaches are often hindered by data noise and the difficulty of capturing fine-grained grammatical nuances. In contrast, diffusion models have recently redefined sign language generation by operating within compressed, structured latent spaces \cite{fang2025stablesignerhierarchicalsign, fang2025signllmsignlanguageproduction}. Departing from existing SLR paradigms, our work introduces \textit{a novel recognition framework that performs inference directly within a sign-specific latent space.} Inspired by the high-efficiency latent modeling in diffusion processes \cite{rombach2021highresolution, fang2025streamflowtheoryalgorithmimplementation}, we leverage a pose-rich latent domain to bypass the limitations of raw data processing. This shift allows our model to more effectively capture complex spatiotemporal dependencies and linguistic structures, establishing a new direction for robust and efficient ASL recognition.

\textbf{Vision Transformer.}
Since the development of ViT \cite{dosovitskiy2020vit}, the use of Transformers in vision tasks has become increasingly widespread. Compared to traditional architectures, Transformers can better model long-range dependencies through self-attention mechanisms \cite{saunders2021continuous, huang2021towards}. The original ViT first applied pure Transformer structure to image classification, pioneering a new paradigm of vision Transformers. Subsequent improvements, such as Swin Transformer and other models, further enhanced model performance in vision tasks by introducing local attention mechanisms \cite{camgoz2020sign, ko2019neural}.
In the field of video event recognition, researchers have introduced spatiotemporal attention mechanisms, enabling ViT to better process video sequence data \cite{Bohacek_2022_WACV, forster2012rwth}. This advance is particularly important for SL recognition, as SL videos contain rich temporal information. Through well-designed spatiotemporal attention mechanisms, models can effectively capture temporal dependencies in  sequences of signing \cite{koller15:cslr, saunders2021mixed}.
\begin{figure*}[t]
  \centering
   \includegraphics[width=0.99\linewidth]{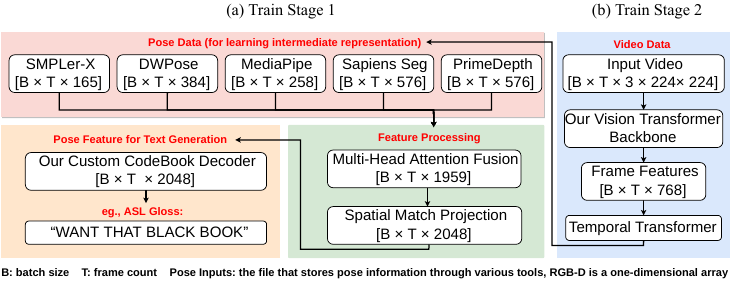}
   \vspace{-6pt}
   \caption{\textbf{Building Compact Pose-Rich Latent Space:} Overall, we utilize ViT \cite{dosovitskiy2020vit} to construct and accommodate a pose latent space. It has two entry points: a video entry at the top layer and a pose data entry in the middle section. \textcolor{purple}{(a) For training stage 1} (Sec. \ref{subsec:multifusion}), we first train the pose fusion layer to output simple text information, ensuring that the learned pose representations are meaningful. \textcolor{purple}{(b) For training stage 2} (Sec. \ref{subsec:Pose_Representations}), we freeze all other components and only learn how RGB videos can be correctly converted into our pose features. \textcolor{purple}{For inference}, only RGB is used as input, so we must ensure that we can encode RGB inputs into our pose features.}
   \label{fig:method}
   \vspace{-4pt}
\end{figure*}

\section{Building Compact Pose-Rich Latent Space}
\label{sec:methodology}

\subsection{Data Processing}

To address the challenge of integrating heterogeneous pose information for SL recognition, we first construct a comprehensive processing pipeline that standardizes multiple pose formats. First, we use a high-quality ASLLRP SignStream® 3 Corpus \cite{neidle2022asl,asllrp2025signbank} dataset, which contains over 80 hours of American Sign Language (ASL) videos with synchronized front view, side view, and facial close-up recordings.
For each input video $V \in \mathbb{R}^{T \times H \times W \times 3}$, where $T$ represents the number of frames and $H,W$ represent height and width, respectively, we extract sequential pose representations for each pose type: $P_{\text{DWPose}} \in \mathbb{R}^{T \times 384}$, $P_{\text{MediaPipe}} \in \mathbb{R}^{T \times 258}$, $P_{\text{SMPLer-X}} \in \mathbb{R}^{T \times 165}$, $P_{\text{PrimeDepth}} \in \mathbb{R}^{T \times 576}$, $P_{\text{Sapiens}} \in \mathbb{R}^{T \times 576}$.
The dimensionality of each pose type reflects its specific information structure. For instance, DWPose contains 18 body keypoints, 21 keypoints for each hand, and 68 facial keypoints, along with their respective confidence scores, totaling 384 dimensions per frame. 

As shown in Fig. \ref{fig:method}, we flatten the information from each frame into a 1-dimensional list, concatenate five types of poses to form an input of length 1959, which is then passed through projection layers for simple transformation and dimension matching; this is then followed by the fusion layer for multimodal integration. 
In this way, we obtain rich pose information for each frame. This standardized structure not only prevents different types of pose information from interfering with each other, but also makes heterogeneous types of pose information mutually compatible.

\subsection{Multimodal Pose Fusion}
\label{subsec:multifusion}

\textbf{In order to ensure that the multi-track pose features are meaningful}, we first conduct a simple training to enable the fused posture features to be easily converted into text:
We modified the pose entry to output layer of ViT approach that learns the mapping from the unified pose representation to natural language descriptions. It consists of several key components: a pose encoder for multimodal fusion, some layers for the feature processing, and a text decoder for generating the final output.

We choose ViT for pose latent space construction because: (1) ViT's self-attention mechanism can effectively fuse heterogeneous pose representations from different extraction tools (SMPLer-X, DWPose, Mediapipe, PrimeDepth, Sapiens) into a unified latent space. (2) ViT's deep layered architecture naturally creates a hierarchical latent space that can compress rich, multi-source pose information into compact representations while preserving essential pose details—making it an ideal container for our pose-rich latent space. (3) Unlike specialized VAE encoders \cite{he2022masked}, ViT's transformer architecture can scale well to high-dimensional inputs, enabling us to replicate Stable Diffusion's success of operating in compact latent space for the sign language domain.

\paragraph{ Multimodal Pose Fusion Layer.}
The pose encoder $E_{\text{pose}}$ processes the five types of pose information with type-specific encoders and fuses them through a multi-head attention mechanism:
\begin{equation}
\begin{aligned}
f_i &= E_i(P_i), \\
f_{\text{fused}} &= \text{MultiHeadAttention}(f_1, f_2, \dots, f_5)
\end{aligned}
\end{equation}
where $E_i$ is the encoder for pose type $i$, and $f_i \in \mathbb{R}^{d_h}$ is the encoded feature. This fusion approach preserves the unique characteristics of each pose type, while allowing information sharing among them \cite{ko2019neural,liu2022bevfusion,zhang2020fusionnet}.

For each $f_i$, we project it to a unified representation space through a dimension matching layer:
\begin{equation}
z_i = \text{PadMatch}(f_i) \in \mathbb{R}^{2048}
\end{equation}
where $z_i$ is the unified representation at position $i$; \text{PadMatch} refers to zero-padding followed by projection. The ViT layers process this unified representation through multi-head self-attention and cross-attention mechanisms:
\begin{equation}
z_{\text{latent}} = \text{ViT}_{\text{layers}}(z_1, z_2, ..., z_n)
\end{equation}
where $z_{\text{latent}}$ represents the final pose-rich latent representation that captures the essence of sign language movements \cite{stoll2020text2sign,cho2021unifying,lu_unified-io_2023}.

\paragraph{Custom CodeBook Decoder.}

We initially attempted to use T5 \cite{2020t5}, currently one of the most popular text decoders, as our text decoding component. However, experimental results revealed that for sign language translation tasks, we only need to consider the vocabulary present in our specific dataset. Large-scale decoders like T5 not only reduce inference speed, but also increase error probability, potentially generating words that do not exist in our target vocabulary $\mathcal{V}_{\text{dataset}}$. Therefore, we developed a custom CodeBook using training data, where the model outputs vocabulary indices rather than raw text tokens, significantly improving accuracy.

\begin{figure*}[t]
  \centering
   \includegraphics[width=0.99\linewidth]{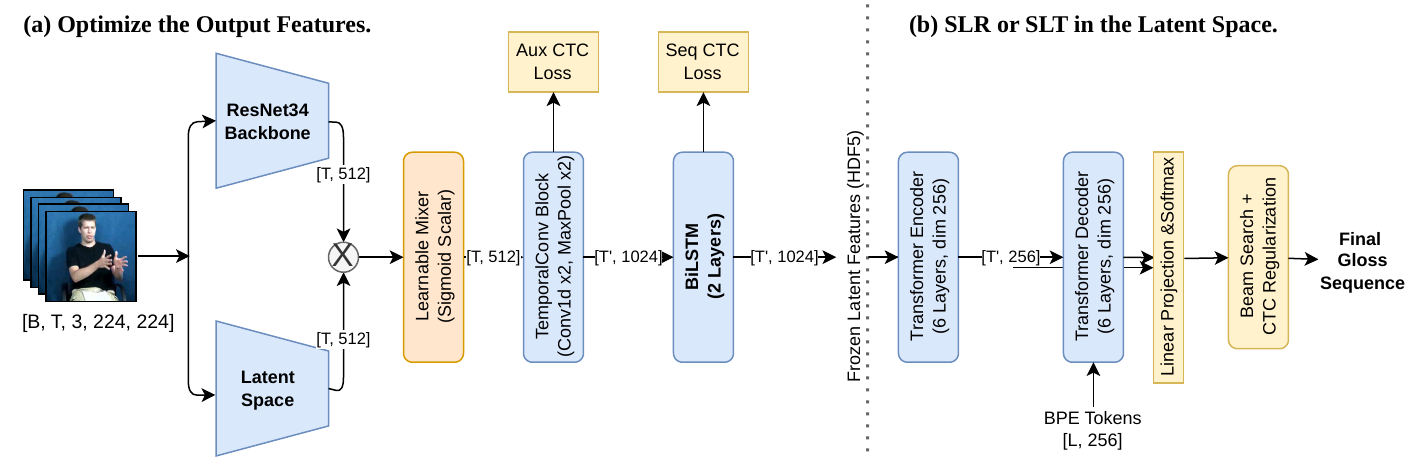}
   \vspace{-10pt}
   \caption{\textbf{Organize and conduct continuous recognition in the latent space:} \textcolor{purple}{(a)} For the latent space of the enriched poses, we further distill, compress and organize it. The number of features trained should be as aligned as possible with the number of Gloss. This can further enhance the upper limit and performance of our model. \textcolor{purple}{(b)} Then, we develop it based on a BiLSTM method \cite{zhang2023sltunet}, enabling it to perform continuous Sign recognition in this latent space, thereby achieving results far superior to previous works.}
   \label{fig:overview}
   \vspace{-8pt}
\end{figure*}

\subsection{Learning Video-to-Pose Representations}
\label{subsec:Pose_Representations}
To process raw video inputs directly, we develop a Video2Pose module based on video entry of ViT \cite{dosovitskiy2020vit}. Given an input video $V \in \mathbb{R}^{T \times H \times W \times 3}$, the Video2Pose module extracts frame-level features and models their temporal relationships:
\begin{equation}
V_{\text{frame}} = {v_1, v_2, ..., v_T} = \text{ViT}_{\text{spatial}}(V)
\end{equation}
where $v_t \in \mathbb{R}^{d_v}$ represents the feature of frame $t$ extracted by the spatial ViT.
To capture temporal dependencies, we employ a temporal attention mechanism:
\begin{equation}
V_{\text{temp}} = \text{TemporalAttention}(V_{\text{frame}})
\end{equation}
The Video2Pose module then projects these temporal features to the same format as the five pose types:
\begin{equation}
\hat{P}_i = \text{Projection}i(V_{\text{temp}})
\end{equation}
where $\hat{P}_i$ is the predicted pose of type $i$. This end-to-end approach eliminates the need for separate pose extraction models during deployment, greatly simplifying the application process \cite{saunders2021mixed,slt-how2sign-wicv2023,chen2020simple,xie2021segformer,xie2021simmim}.
\section{Continuous Sign Language Recognition in the Latent Space}

In the previous section, we described the construction of a latent space based on ViT and through simple training, it was able to convert the original video into meaningful representations. Next, we should develop an identification method that is trained in the latent space.

  \subsection{Latent Feature Organization and Optimization}

\begin{wrapfigure}{r}{0.48\columnwidth}
  \vspace{-22pt}
  \centering
  \footnotesize
  \noindent\rule{\linewidth}{0.4pt}
  \vspace{-8pt}
  {\setlength{\abovecaptionskip}{0pt}
   \setlength{\belowcaptionskip}{0pt}
   \captionsetup{type=algorithm}
   \captionsetup{type=algorithm, justification=raggedright, singlelinecheck=false}
   \captionof{algorithm}{PoseAwareFeatureCompilation}
   \label{alg:pose_compile}}
  \vspace{-8pt}
  \noindent\rule{\linewidth}{0.4pt}
  {\setlength{\topsep}{0pt}
  \begin{algorithmic}[1]
  \vspace{-2pt}
    \Procedure{Compile}{$\{z_t, g_t, m_t\}_{t=1}^{T}$}
      \State $z_t \gets \text{LayerNorm}(z_t)$
      \State $c_t \gets \text{CrossCov}(z_t, z_{t-1})$
      \State $z_t \gets z_t - \gamma \cdot c_t$ \Comment{Covariance whitening}
      \If{$\text{Bernoulli}(\rho) = 1$}
        \State $z_t \gets \mathbf{0}$; $m_t \gets 0$ \Comment{Stochastic frame drop}
      \EndIf
      \State $a_t \gets \text{GlossAlign}(z_t, g_t)$
      \State \Return $\{(z_t, a_t, m_t)\}_{t=1}^{T}$
    \EndProcedure
  \end{algorithmic}}
  \vspace{-8pt}
  \noindent\rule{\linewidth}{0.4pt}
  \vspace{-32pt}
\end{wrapfigure}

  \paragraph{Pose-Aware Feature Compilation.}
  Before the temporal learner can reason over sequences, we enforce a disciplined organization of the features produced by the Vid2Pose encoder. Every frame-level embedding $z_t \in \mathbb{R}^{2048}$ is accompanied by its gloss supervision signal $g_t$ and confidence mask $m_t$, giving us a richer tuple $\langle z_t, g_t, m_t\rangle$. As shown in Fig. \ref{fig:overview}, this tuple is passed through a \textit{lightweight transformer stack} that performs \textcolor{purple}{(i)} variance-preserving layer normalization, \textcolor{purple}{(ii)} cross-frame covariance whitening (using coefficient $\gamma$), and \textcolor{purple}{(iii)} stochastic frame dropping (with probability $\rho$) to avoid overfitting to repetitive segments. The process aligns pose cues with textual semantics via a \textit{GlossAlign} function 
  this process is formalized in Algorithm \ref{alg:pose_compile}. Here, GlossAlign assigns each feature to its corresponding gloss based on temporal overlap, and CrossCov \cite{zhang2023sltunet} computes the cross-covariance between consecutive frames for redundancy detection.

  \paragraph{Gloss-Aware Latent Distillation.}
  With the compiled tuples we run a \textit{knowledge distillation} objective that pushes the latent vectors toward a compact space guided by gloss embeddings. A teacher backbone $f_{\text{ResNet}}$ (based on ResNet-34) produces logits $y_t^{\text{teacher}}$, while the latent vectors generate student logits $y_t^{\text{student}}$ through a two-layer perceptron. The KL divergence between both streams, weighted by gloss frequency priors, ensures that rare glosses receive more attention. This mirrors the \textit{SeqKD} objective but operates entirely within our latent pose space. Algorithm \ref{alg:latent_train} presents the complete hierarchical training procedure.

\paragraph{Adaptive Feature Pruning.} 
To optimize the information density of the latent space, we monitor each latent dimension's importance during training using its \textit{Fisher information}. Specifically, dimensions whose activation variance fails to surpass a dynamic threshold $\tau$ are identified as redundant and subsequently pruned:
\begin{equation}
\text{Prune } z^{(d)} \text{ if } Var(z^{(d)}) < \tau
\end{equation}
This mechanism reduces the effective latent width to approximately 1024, which forces the model to grasp the similarities and differences between different pose formats. To ensure structural compatibility with the subsequent BiLSTM and Transformer layers without altering tensor shapes, we implement this as a masking operation:
\begin{equation}
\hat{z}_t = z_t \odot M, \quad M_d = 
\begin{cases} 
1 & \text{if } Var(z^{(d)}) \geq \tau \\
0 & \text{otherwise}
\end{cases}
\end{equation}
This pruning step is triggered every $K$ iterations and is synchronized with the checkpoint averaging routine to maintain stability during the transition.

\begin{wrapfigure}{r}{0.48\columnwidth}
  \vspace{-22pt}
  \centering
  \footnotesize
  \noindent\rule{\linewidth}{0.4pt}
  \vspace{-8pt}
  {\setlength{\abovecaptionskip}{0pt}
   \setlength{\belowcaptionskip}{0pt}
   \captionsetup{type=algorithm}
   \captionsetup{type=algorithm, justification=raggedright, singlelinecheck=false}
   \captionof{algorithm}{HierarchicalLatentTraining}
   \label{alg:latent_train}}
  \vspace{-8pt}
  \noindent\rule{\linewidth}{0.4pt}
  {\setlength{\topsep}{0pt}
  \begin{algorithmic}[1]
  \vspace{-2pt}
    \Procedure{TrainLatent}{$\mathcal{D}$}
      \For{epoch $= 1 \ldots E$}
        \For{batch $\in \mathcal{D}$}
          \State $\{(z_t, a_t, m_t)\} \gets \textsc{Compile}(\text{batch})$
          \State $y^{\text{teacher}} \gets f_{\text{ResNet}}(\text{batch})$
          \State $y^{\text{student}} \gets f_{\text{MLP}}(\{z_t\})$
          \State $\mathcal{L}_{\text{KD}} \gets \text{KL}(y^{\text{student}}, y^{\text{teacher}})$
          \State $\mathcal{L}_{\text{CTC}} \gets \text{CTC}(y^{\text{student}}, \text{gloss})$
          \State $\mathcal{L} \gets \lambda_{\text{KD}}\mathcal{L}_{\text{KD}} + \lambda_{\text{CTC}}\mathcal{L}_{\text{CTC}}$
          \State Update params via Adam
        \EndFor
        \If{$\text{epoch} \bmod K = 0$}
          \State Prune dims: $\text{Var}(z^{(d)}) < \tau$
        \EndIf
      \EndFor
    \EndProcedure
  \end{algorithmic}}
  \vspace{-8pt}
  \noindent\rule{\linewidth}{0.4pt}
  \vspace{-32pt}
\end{wrapfigure}

\paragraph{Multi-Scale Feature Augmentation.} To combat data scarcity in sign language datasets, we apply a $N=10$-fold temporal augmentation during feature extraction \cite{zhang2023sltunet}. For each video sample $i$, the SMKD model \cite{zhang2023sltunet} generates $N$ augmented versions through (i) random temporal scaling $\sigma \in [0.8, 1.2]$, (ii) spatial jittering with probability $p_{jit}=0.3$, and (iii) feature-space Gaussian noise $\mathcal{N}(0, 0.01)$. The augmented features are stored in HDF5 format with keys $\{i\}_{j}$ where $j \in [0, N-1]$ indexes the augmentation version. It enables efficient random access during training while expanding the effective data size from $M$ to $M \times N$ samples. Unlike pixel-space augmentation which requires re-encoding, our latent augmentation operates on extracted features, reducing computational overhead.

  \subsection{Latent-Space Temporal Modeling and Decoding}

  \paragraph{Hybrid Temporal Modeling.}
  The distilled latent sequences feed a hybrid backbone that remains modality-agnostic. First, a \textit{TemporalConv stack} (two K5 blocks interleaved with stride-2 pooling) compresses sequences from $T$ frames to $T' \approx T/4$, producing features $h_t \in \mathbb{R}^{1024}$. These features then pass through a \textit{bidirectional LSTM} (2 layers, hidden size 512 per direction) to capture long-range dependencies, yielding $u_t \in \mathbb{R}^{1024}$. Because all computations remain within the latent space, we avoid heavy pixel-level operations during inference.

\begin{wrapfigure}{r}{0.48\columnwidth}
  \centering
  \footnotesize
  \noindent\rule{\linewidth}{0.4pt}
  \vspace{-8pt}
  {\setlength{\abovecaptionskip}{0pt}
   \setlength{\belowcaptionskip}{0pt}
   \captionsetup{type=algorithm}
   \captionsetup{type=algorithm, justification=raggedright, singlelinecheck=false}
   \captionof{algorithm}{LatentSequenceDecoding}
   \label{alg:latent_decode}}
  \vspace{-8pt}
  \noindent\rule{\linewidth}{0.4pt}
  {\setlength{\topsep}{0pt}
  \begin{algorithmic}[1]
  \vspace{-2pt}
    \Procedure{Decode}{$H{=}\{u_t\}_{t=1}^{T'}$, beam $B$}
      \State Init $\mathcal{B} \gets \{\langle \text{BOS}, 0\rangle\}$
      \For{$t = 1 \ldots T_{\max}$}
        \State $\mathcal{B}' \gets \emptyset$
        \For{hyp $\langle y, s\rangle \in \mathcal{B}$}
          \State $p(\cdot) \gets \text{DecoderStep}(y, H)$
          \For{$w$ in top-$k$ of $p$}
            \State $s' \gets s + \log p(w)$
            \State $\quad - \alpha \cdot \text{AttnEntropy}(y, H)$
            \State $\mathcal{B}' \gets \mathcal{B}' \cup \{\langle y{\oplus}w, s'\rangle\}$
          \EndFor
        \EndFor
        \State $\mathcal{B} \gets$ top-$B$ of $\mathcal{B}'$
        \If{all beams end with EOS} \textbf{break} \EndIf
      \EndFor
      \State \Return best hyp in $\mathcal{B}$
    \EndProcedure
  \end{algorithmic}}
  \vspace{-8pt}
  \noindent\rule{\linewidth}{0.4pt}
  \vspace{-36pt}
\end{wrapfigure}

  \paragraph{CTC-Constrained Transformer Refinement.}
  A \textit{Transformer encoder-decoder} (6 layers each, width 256) refines the BiLSTM outputs into gloss sequences. The encoder ingests $u_t$ after projection to 256 dimensions, whereas the decoder consumes BPE-tokenized gloss strings. We attach a \textit{CTC head} directly to the encoder outputs to enforce monotonic alignment, ensuring the architecture accepts the latent tensors without additional reshaping. The joint objective combines sequence-level cross-entropy, encoder-side CTC, and a \textit{latent-space regularizer} that encourages successive encoder states to remain within a Lipschitz ball.

  \paragraph{Beam Search with Latent Feedback.}
  During inference we run an 8-beam search where each hypothesis carries an auxiliary latent score derived from the \textit{encoder attention entropy}. Hypotheses with erratic attention (high entropy) incur a penalty $\alpha$, biasing the search toward sequences that remain faithful to the latent pose evidence. The final gloss output can therefore be obtained without re-running pose estimators. Algorithm \ref{alg:latent_decode} details the complete beam search decoding procedure.

    \paragraph{Training Optimization and Regularization.}
  To stabilize convergence and improve generalization, we employ a suite of
  optimization techniques. The learning rate follows a Noam
  scheduler~\cite{vaswani2017attention} with warmup steps $W=4000$:
  \begin{equation}
  \eta(t) = d_{model}^{-0.5} \cdot \min(t^{-0.5}, t \cdot W^{-1.5})
  \end{equation}
  where $d_{model}=256$. We apply \textit{differentiated dropout} with rates
  tailored to each component: $p_{attn}=0.3$ for attention weights,
  $p_{relu}=0.5$ for activation layers, and $p_{res}=0.4$ for residual
  connections, preventing over-reliance on any single pathway. Label smoothing
   with $\epsilon=0.1$ regularizes the output distribution.
  Finally, we adopt \textit{checkpoint averaging} over the top-5 models ranked by
  validation WER.
\begin{table*}[t]   
  \centering
  \setlength\tabcolsep{3pt}

    \resizebox{0.98\textwidth}{!}{
  \begin{tabular}{cl|ccccc|ccccc}
    \hline 
    \hline
    \multicolumn{12}{c}{ ASLLRP } \\
    \hline 
    & \multirow{2}{*}{Method}& \multicolumn{5}{c|}{ Dev } & \multicolumn{5}{c}{ Test }  \\
    & &Rouge & BLEU1 & BLEU2 & BLEU3 & BLEU4 & Rouge & BLEU1 &BLEU2 & BLEU3 & BLEU4 \\
    \hline 
    \multirow{4}{*}{ Sign2Gloss }
    & C2SLR~\cite{zuo2022c2slr} & 52.50 & 50.10 & 38.20 & 29.50 & 24.76 & 52.30 & 49.95 & 38.00 & 29.35 & 24.58 \\
    & CorrNet+~\cite{hu2024corrnetsignlanguagerecognition} & 53.80 & 51.40 & 39.50 & 31.20 & 26.14 & 53.60 & 51.25 & 39.30 & 31.05 & 25.96\\
    & SLTUNET~\cite{zhang2023sltunet} & 54.20 & 52.10 & 40.10 & 31.80 & 26.48 & 54.05 & 51.90 & 39.95 & 31.65 & 26.31 \\
    & \textbf{SignX (Ours)} & \textbf{56.65} & \textbf{51.84} & \textbf{42.48} & \textbf{35.35} & \textbf{29.80} & \textbf{56.48} & \textbf{51.65} & \textbf{42.31} & \textbf{35.19} & \textbf{29.64}\\
    \hline
    \end{tabular}
    }\vspace{6pt}
      \caption{\textbf{Benchmark on the ASLLRP dataset}~\cite{neidle2022asl,asllrp2025signbank}. To ensure a fair and rigorous comparison on the ASLLRP dataset, we re-implemented and fine-tuned several representative SLR frameworks, including C2SLR and SLTUNET, achieving their best possible performance as modern baselines.
      } 
    \label{tabe:sota_bleu}
\end{table*}

\begin{table}[ht]
\begin{minipage}{0.5\columnwidth}
  \centering
  \vspace{-12pt}
  \scriptsize
  \newcolumntype{P}{>{\RaggedRight\arraybackslash}p{1\linewidth}}
  \begin{tabular}{P}
  \toprule
  \textbf{SignX (Ours)} \\
  \textit{Generated Text:} IF STUDENT IX HEAR/LISTEN TEACH+AGENT FUTURE LEARN SOMETHING/ONE \\
  \textit{Ground Truth:} IF STUDENT IX HEAR/LISTEN TEACH+AGENT FUTURE LEARN SOMETHING/ONE \\
  \cmidrule{1-1}
  \textit{Generated Text:} BOX/ROOM IX NOT-YET ARRIVE IX SHOULD CONTACT ns-fs-FEDEX \\
  \textit{Ground Truth:} BOX/ROOM IX NOT-YET ARRIVE IX SHOULD CONTACT ns-fs-FEDEX \\
  \cmidrule{1-1}
  \textit{Generated Text:} PEOPLE ARRIVE LATE WHO \\
  \textit{Ground Truth:} PERSON ARRIVE LATE WHO \\
  \cmidrule{1-1}
  \textit{Generated Text:} HERE ns-fs-MARY FUTURE PAY/BUY CAR \\
  \textit{Ground Truth:} MAYBE ns-fs-MARY FUTURE PAY/BUY CAR \\
  \bottomrule
  \end{tabular}
  \vspace{6pt}
  \caption{\textbf{Qualitative evaluation of translation results on the ASLLRP dataset.} The examples demonstrate successful reconstructions of long-form gloss sequences and specialized fingerspelling, as well as minor failure cases. Other examples can be found in Appendix Section~\ref{subsec:qualitative}.}
  \vspace{-12pt}
  \label{tab:qualitative_short}
\end{minipage}\hspace*{0.02\columnwidth}%
\begin{minipage}{0.5\columnwidth}
  \centering
  \vspace{-8pt}
  \begin{subfigure}{0.485\linewidth}
    \centering
    \includegraphics[width=\linewidth]{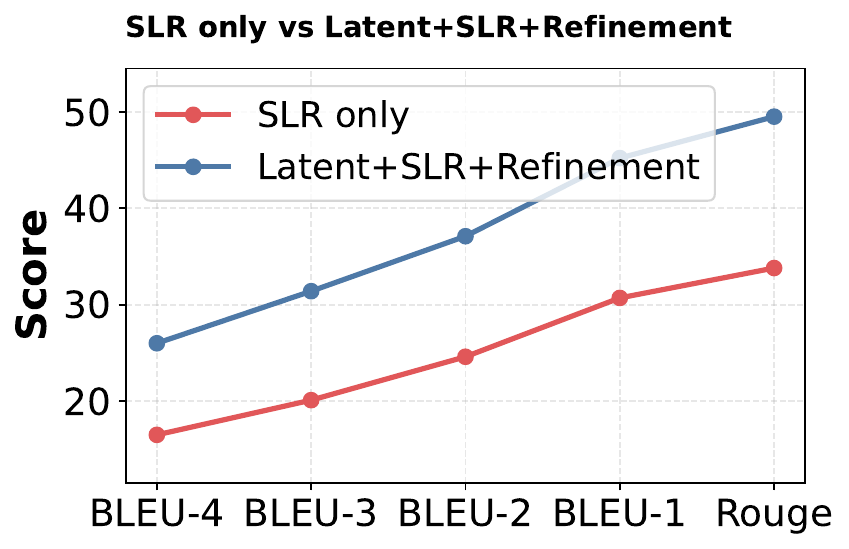}
    \caption{SLR only vs Full}
  \end{subfigure}\hfill
  \begin{subfigure}{0.485\linewidth}
    \centering
    \includegraphics[width=\linewidth]{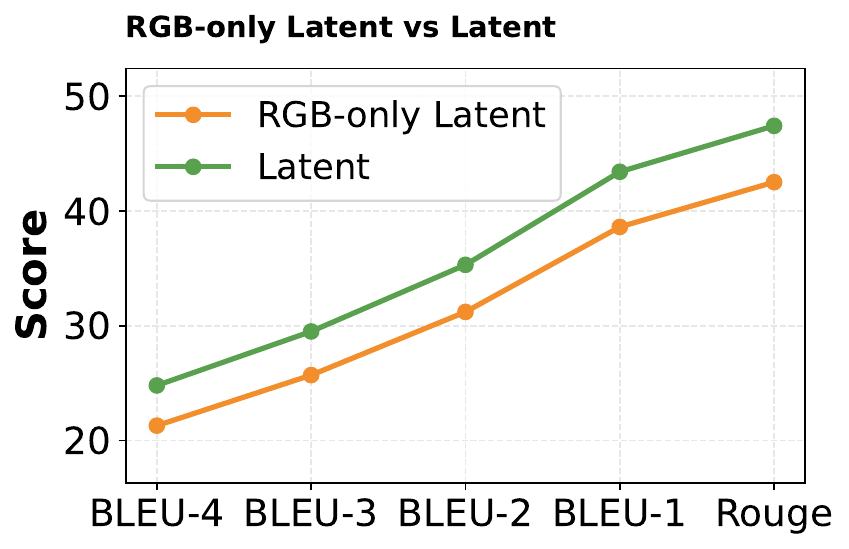}
    \caption{RGB-only vs Pose}
  \end{subfigure}
  \vspace{4pt}
  \begin{subfigure}{0.485\linewidth}
    \centering
    \includegraphics[width=\linewidth]{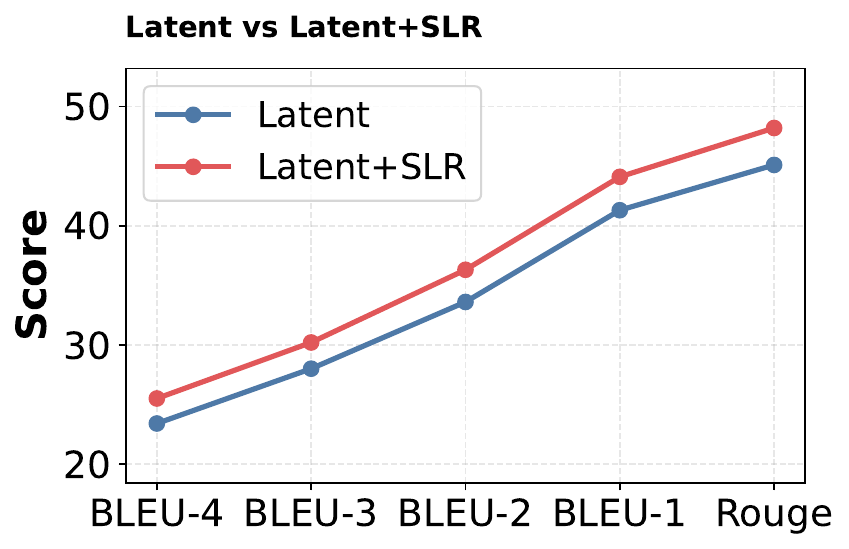}
    \caption{Latent vs +SLR}
  \end{subfigure}\hfill
  \begin{subfigure}{0.485\linewidth}
    \centering
    \includegraphics[width=\linewidth]{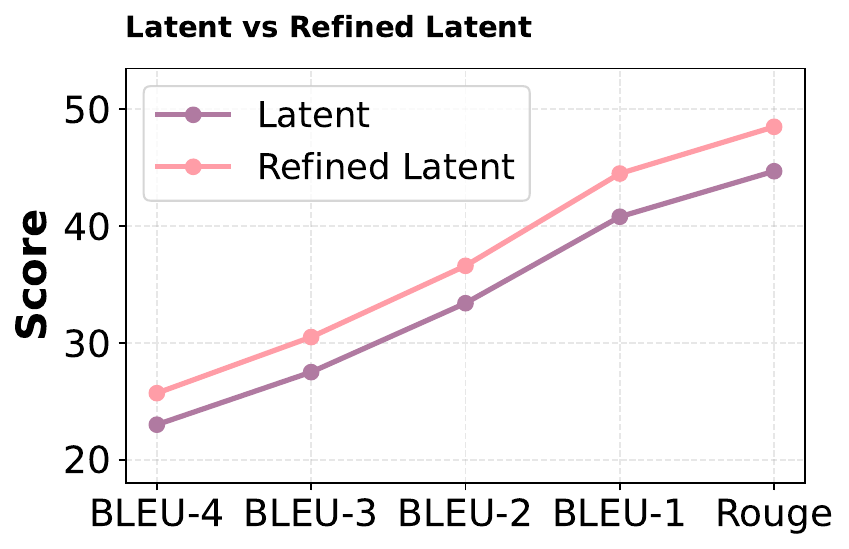}
    \caption{Latent vs Refined}
  \end{subfigure}
  \vspace{-4pt}
  \captionof{figure}{\textbf{Ablation Study Results.} Performance comparison across different 1k step model configurations on ASLLRP dev set. (a) Impact of full pipeline components. (b) Effect of pose feature integration. (c) Contribution of sign language recognition module. (d) Improvement from latent refinement. More results are available at Sec.~\ref{subsec:Abl_of_Latent_Space} and Appendix Sec.~\ref{subsec:Training_Efficiency_of_Latent_Space}.}
  \label{fig:ablation_study}
\end{minipage}
\vspace{-12pt}
\end{table}

\section{Experimental Results}
\label{sec:results}

Here, we evaluate our SignX continuous recognition method on mainstream datasets (ASLLRP \cite{asllrp2025signbank}: 1,686/210/212 train/dev/test; PHOENIX2014-T \cite{camgoz2018neural}: 7,096/519/642 train/val/test; CSL-Daily \cite{zhou2021improving}: 18,401/1,077/1,176; WLASL-2000 \cite{WLASL}: 8,618/479/478 train/val/test), using metrics such as Bleu \cite{bleu}, WER \cite{KLAKOW200219}, and P-I \cite{li2020word}.
We conduct a series of experiments including qualitative, quantitative, ablation, and efficiency evaluations, demonstrating that our method has achieved a significant improvement over the state-of-the-art\footnote{More mertric information in Appendix Section \ref{subsec:more_metric}.}.


\subsection{Qualitative Assessment}

SignX consistently reproduces sentence-level semantics on long-form conditionals (Table \ref{tab:qualitative_short}, first row), suggesting that the hierarchical design effectively preserves subject--temporal relations over extended contexts. The second sample highlights a specialized strength in accurately capturing fingerspelled entities like ``fs-FEDEX'' within domain-specific logistics, maintaining the structural framing and lexical integrity entirely intact. The third sample demonstrates the model's capacity for semantic alignment; although the lexical choice of ``PEOPLE'' diverges from the ground-truth ``PERSON,'' the core agentive meaning is successfully recovered through latent synonym mapping. The final example exhibits a minor stylistic drift where the modal adverb ``MAYBE'' is substituted with ``HERE''---a frequent spatial filler in sign language---reflecting a slight divergence in functional modifiers rather than a structural collapse. Overall, the model captures core semantic intent with high fidelity but may substitute ground-truth glosses with semantically refined or nuanced alternatives.


\begin{table*}[t]   
  \centering
  \setlength\tabcolsep{3pt}

    \resizebox{0.98\textwidth}{!}{
  \begin{tabular}{cl|ccccc|ccccc}
    \hline 
    \hline
    \multicolumn{12}{c}{ PHOENIX2014-T } \\
    \hline 
    & \multirow{2}{*}{Method}& \multicolumn{5}{c|}{ Dev } & \multicolumn{5}{c}{ Test }  \\
    & &Rouge & BLEU1 & BLEU2 & BLEU3 & BLEU4 & Rouge & BLEU1 &BLEU2 & BLEU3 & BLEU4 \\
    \hline 
    \multirow{8}{*}{ Sign2Text }&  SL-Luong~\cite{camgoz2018neural} & 31.80 & 31.87 & 19.11 & 13.16 & 9.94 & 31.80 & 32.24 & 19.03 & 12.83 & 9.58 \\
    & Joint-SLRT~\cite{camgoz2020sign} &- & 47.26 & 34.40 & 27.05 & 22.38 & - & 46.61 & 33.73 & 26.19 & 21.32 \\
    & STMC-T~\cite{zhou2021spatial} & 48.24 & 47.60 & 36.43 & 29.18 & 24.09 & 46.65 & 46.98 & 36.09 & 28.70 & 23.65 \\
    & SignBT~\cite{zhou2021improving}& 50.29 & 51.11 & 37.90 & 29.80 & 24.45 & 49.54 & 50.80 & 37.75 & 29.72 & 24.32 \\
    & MMTLB~\cite{chen2022simple}& 53.10 & 53.95 & 41.12 & 33.14 & 27.61 & 52.65 & 53.97 & 41.75 & 33.84 & 28.39 \\
    & SLTUNET~\cite{zhang2023sltunet} & 52.23 & - & - & - & 27.87 & 52.11 & 52.92 & 41.76 & 33.99 & 28.47\\
    & TwoStream-SLT~\cite{chen2022two} & 54.08 & 54.32 & 41.99 & 34.15 & 28.66 & 53.48 & 54.90 & 42.43 & 34.46 & 28.95 \\
    & CorrNet+ \cite{hu2024corrnetsignlanguagerecognition} & 54.54 & 54.56 & 42.31 & 34.48 & 29.13 & 53.76 & 55.32 & 42.74 & 34.86 & 29.42\\
    & \textbf{SignX (Ours)} & \textbf{55.32} & \textbf{55.24} & \textbf{43.10} & \textbf{35.45} & \textbf{30.08} & \textbf{55.15} & \textbf{55.08} & \textbf{42.92} & \textbf{35.28} & \textbf{29.91}\\
    \hline
    \hline
    \multicolumn{12}{c}{ CSL-Daily } \\
    \hline 
    & \multirow{2}{*}{Method}& \multicolumn{5}{c|}{ Dev } & \multicolumn{5}{c}{ Test }  \\
    & &Rouge & BLEU1 & BLEU2 & BLEU3 & BLEU4 & Rouge & BLEU1 &BLEU2 & BLEU3 & BLEU4 \\
    \hline 
    \multirow{8}{*}{ Sign2Text }&  SL-Luong~\cite{camgoz2018neural} & 34.28 & 34.22 & 19.72 & 12.24 & 7.96 & 34.54 & 34.16 & 19.57 & 11.84 & 7.56 \\
    & SignBT~\cite{zhou2021improving} & 49.49& 51.46 & 37.23 & 27.51 & 20.80 & 49.31 & 51.42 & 37.26 & 27.76 & 21.34 \\
    & MMTLB~\cite{chen2022simple} & 53.38 & 53.81 & 40.84 & 31.29 & 24.42 & 53.25 & 53.31 & 40.41 & 30.87 & 23.92 \\
    & SLTUNET~\cite{zhang2023sltunet} & 53.58 & - & -& - &23.99 & 54.08 & 54.98 & 41.44 & 31.84 & 25.01 \\
    & TwoStream-SLT~\cite{chen2022two} & 55.10 & 55.21 & 42.31 & 32.71 & 25.76 & 55.72 & 55.44 & 42.59 & 32.87 & 25.79 \\
    & CorrNet+ \cite{hu2024corrnetsignlanguagerecognition} & 55.52 & 55.64 & 42.78 & 33.13 & 26.14 & 55.84 & 55.82 & 42.96 & 33.26 & 26.14\\
    & Uni-Sign \cite{li2025uni} & 56.03 & 55.30 & - & - & 26.25 & 56.51 & 55.08 & - & - & 26.36 \\
    & \textbf{SignX (Ours)} & \textbf{57.25} & \textbf{57.10} & \textbf{44.50} & \textbf{35.80} & \textbf{28.75} & \textbf{57.12} & \textbf{56.95} & \textbf{44.32} & \textbf{35.62} & \textbf{28.58}\\
    \hline
    \end{tabular}
    }
    \vspace{6pt}
      \caption{\textbf{Comparison with state-of-the-art methods} on the PHOENIX2014-T dataset~\cite{camgoz2018neural} and CSL-Daily dataset~\cite{zhou2021improving} over the SLT setting. ``-'': Information is unavailable.} 
    \label{table:sota_bleu}
    \vspace{-12pt}
  \end{table*}

\subsection{Quantitative Assessment}


\subsubsection{Ablation Results}
As illustrated in Fig.~\ref{fig:ablation_study}, ablation studies on ASLLRP confirm that our full pipeline significantly outperforms the SLR-only baseline. Specifically, pose features provide critical spatial cues, while the SLR module and latent refinement mechanism enhance intermediate supervision and temporal coherence, validating each component's necessity in capturing complex sign language semantics. All experiments exhibit consistent improvement trends across metrics, collectively validating each component's necessity in our framework.


\subsection{Comparison with Previous Work}
We evaluate the \textbf{SignX} framework across multiple mainstream benchmarks \cite{8578910, Zhou2021ImprovingSLT-with-monolingual-CSLDaily, li2020word}, utilizing a comprehensive suite of metrics including WER,
\begin{wraptable}{l}{0.49\columnwidth}
\centering
\scriptsize
\resizebox{\linewidth}{!}{%
\begin{tabular}{lccclc}
\toprule
\multicolumn{3}{c}{\textbf{Continuous SLR (CSLR)}} & & \multicolumn{2}{c}{\textbf{Isolated SLR (ISLR)}} \\
\cmidrule(r){1-3} \cmidrule(l){5-6}
Method & RWTH-14T & CSL-Daily & & Method & WLASL-2000 \\
& (WER $\downarrow$) & (WER $\downarrow$) & & & (P-I $\uparrow$) \\
\midrule
GEU \cite{9556136}  & 49.9 & - & & SignBERT \cite{hu2021signbert} & 39.40 \\
SLT \cite{SLT}  & 24.5 & 32.0 & & BEST \cite{zhao2023best} & 46.25 \\
Bert \cite{10542663} & 21.2 & 30.8 & & SignBERT+ \cite{hu2023signbert+} & 48.85 \\
Bert fine-tuning \cite{10542663} & 20.4 & 30.0 & & MSLU \cite{zhou2024scaling} & 56.29 \\
CorrNet \cite{hu2023continuous}  & 20.5 & 30.1 & & NLA-SLR \cite{zuo2023natural} & 61.05 \\
GPGN \cite{10542663} & 20.5 & 30.0 & & Uni-Sign \cite{li2025uni} & 63.52 \\
CorrNet+ \cite{hu2024corrnetsignlanguagerecognition} & 19.1 & 28.2 & & Sigma \cite{pu2025sigmasemanticallyinformativepretraining} & 64.40 \\
\midrule
\textbf{SignX (Ours)} & \textbf{18.6} & \textbf{24.3} & & \textbf{SignX (Ours)} & \textbf{68.29} \\
\bottomrule
\end{tabular}%
}
\caption{\textbf{Comprehensive Comparison on Continuous and Isolated SLR.} Left: WER on RWTH-2014-T and CSL-Daily. Right: P-I accuracy on WLASL-2000. Our SignX framework achieves SOTA performance across all tracks, even when compared with the latest Signma \cite{pu2025sigmasemanticallyinformativepretraining} from the same period.}
\label{tab:performance_v5}
\vspace{-20pt}
\end{wraptable}
BLEU, and P-I accuracy. 
As illustrated in \cref{table:sota_bleu,tab:performance_v5}, SignX achieves significant improvements over current state-of-the-art methods in both continuous sign recognition and isolated sign recognition tasks (Note that some of the SLT work is different from ours and cannot be directly compared, e.g., \cite{zhang2025largesignlanguagemodels}). Overall, our framework establishes a new performance ceiling for both continuous sign language recognition (CSLR) and sign language translation (SLT) tasks, exhibiting superior robustness and cross-domain scalability, thereby providing a high-performance baseline for the broader sign language understanding community.




\begin{wrapfigure}{l}{0.5\textwidth}
  \vspace{-24pt}
  \centering
  \includegraphics[width=0.99\linewidth]{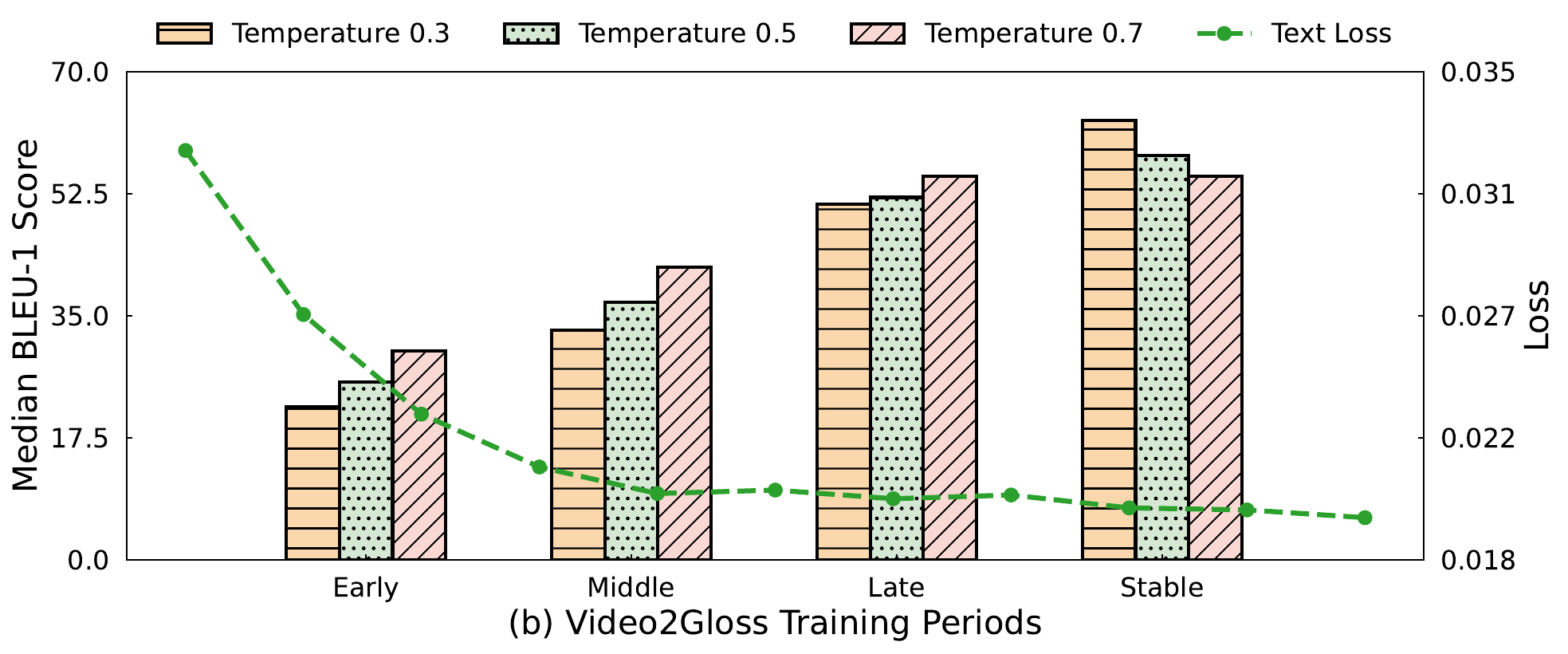}
  \vspace{-18pt}
  \caption{\textbf{Impact of hyperparameter settings on latent space quality.} We evaluate how temperature variations affect the discriminative power of the constructed latent space across different optimization periods. The results highlight the necessity of balancing generation precision and representation stability to maintain a compact pose-rich encoding.}
  \label{fig:latent_representation_quality}
  \vspace{-18pt}
\end{wrapfigure}

\subsection{Ablation Study of Building Compact Pose-Rich Latent Space}
\label{subsec:Abl_of_Latent_Space}
To further refine the properties of the constructed latent space, we conducted an ablation study on the temperature settings, as shown in Fig.~\ref{fig:latent_representation_quality}. We examined the BLEU scores across four temperature levels (0.1, 0.3, 0.5, and 0.7) to determine the optimal configuration for latent space stability and gloss accuracy. The evaluation across early to stable periods reveals that a moderate temperature of 0.3 facilitates the most robust latent representation for sign recognition. We observed that while initial space formation benefits from this moderate setting, the subsequent refinement of the video-to-latent mapping favors lower temperatures to enhance precision. This balance is crucial for ensuring that the latent space remains both expressive and discriminative. Through this optimization, we also established the final weighting scheme for the latent space loss functions.


\begin{table}[ht]
\begin{minipage}{0.49\columnwidth}
  \centering
  \includegraphics[width=\linewidth]{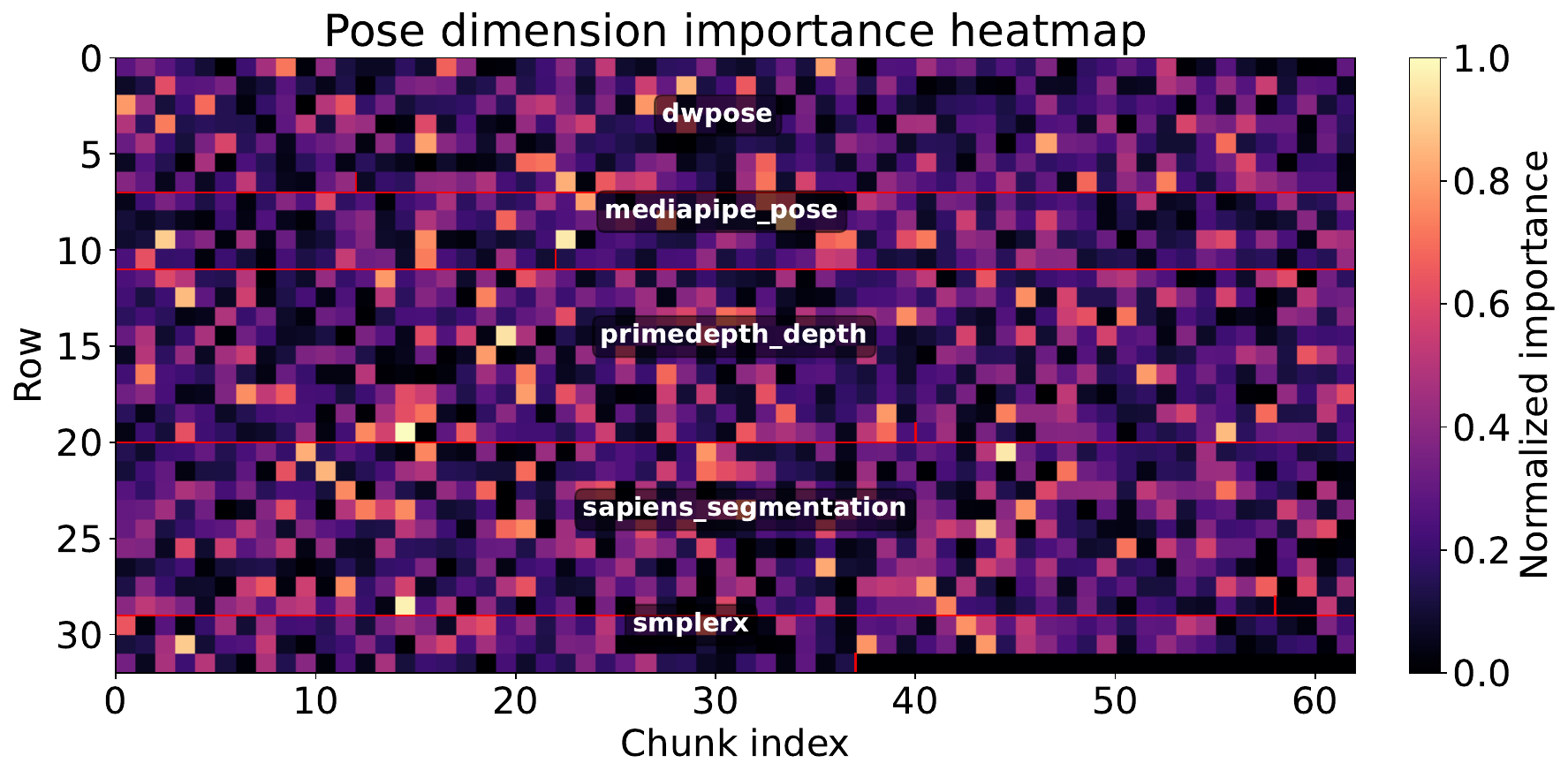}
  \vspace{-7pt}
  \captionof{figure}{\textbf{Importance Visualization of different modalities.} Normalized importance heatmap of 1959 pose feature dimensions across five pose types, separated by red boundary lines. Warmer colors indicate higher activation importance.}
  \label{fig:pose_dimension_heatmap}
  \vspace{-4pt}
\end{minipage}\hspace*{0.02\columnwidth}%
\begin{minipage}{0.49\columnwidth}
  \centering
  \vspace{4pt}
  \scriptsize
  \resizebox{\linewidth}{!}{
  \begin{tabular}{lccc}
  \toprule
  \makecell{\textbf{Ablated Modality} \\ \textbf{(Zeroed)}} & \textbf{WER (\%)} & \textbf{$\Delta$ WER} & \makecell{\textbf{Relative} \\ \textbf{Degradation}} \\ \midrule
  SignX (Full Pipeline)             & 26.59             & --                    & --                            \\
  w/o SMPLer-X                      & 29.65             & +3.06                 & 11.51\%                       \\
  w/o DWPose                        & 32.31             & +5.72                 & 21.51\%                       \\
  w/o Mediapipe                     & 29.37             & +2.78                 & 10.46\%                       \\
  w/o PrimeDepth                    & 30.89             & +4.30                 & 16.17\%                       \\
  w/o Sapiens                       & 30.53             & +3.94                 & 14.82\%                       \\ \bottomrule
  \end{tabular}
  }
  \vspace{9pt}
  \caption{\textbf{Ablation assessment of different modalities:} For each modality mentioned in the first column, the corresponding output is set to zero during video processing. Recognition is then performed within the corrupted latent space to observe the impact on model performance.}
  \label{tab:modality_ablation_final}
\end{minipage}
\vspace{-24pt}
\end{table}

\subsection{Ablation Assessment of Five modalities}

As shown in Figure \ref{fig:pose_dimension_heatmap}, during the extraction process of Video2Pose, we analyzed the attention intensities of different dimensions. We can observe that the intensities of different dimensions of the ViT used for accommodating the creation of the Latent Space after training are very evenly distributed. Therefore, we can conclude that during the main stage where it plays a role, all the postures have similar levels of importance.

Table \ref{tab:modality_ablation_final} summarizes a further, more granular assessment of modality ablation, in which specific modality outputs are zeroed to generate various corrupted versions of the latent space for a detailed analysis of each modality's contribution to the overall framework. The Full Pipeline model is a Vid2Pose module trained for 30 epochs. Upon zeroing a modality, the CSLR model is fine-tuned for 1,000 steps within the resulting corrupted latent space before performance comparison. We perform restorative training within these corrupted spaces to observe the impact of missing modalities on downstream recognition, moving beyond the evaluation of the initial extraction pipeline. As illustrated in the table, DWPose and depth information exert the most significant influence on performance, which aligns with the conventional consensus that these two types of data are paramount for sign language recognition. The lowest impact is observed from MediaPipe Pose, which we hypothesize is because its functional role is largely subsumed by the superior precision of DWPose. Notably, although we performed 1,000 steps of additional restorative fine-tuning for each configuration, the aggregate performance degradation remains relatively controlled. This phenomenon indicates that a substantial portion of linguistic information is shared across different modalities, thereby validating our original motivation for constructing a compact and highly compressed latent space.

\subsection{Inference Efficiency Assessment}

Table~\ref{tab:efficiency} presents a comparative analysis of inference efficiency and recognition performance on the ASLLRP dataset. The results demonstrate the clear superiority of the proposed SignX (CSLR/SLT in Latent Space) across all evaluated metrics. Notably, the latent space variant achieves an inference speed of \texttt{2.42 FPS}, representing an approximately 4.2$\times$ speedup over its pixel-space counterpart (0.57 FPS) and nearly a 50-fold acceleration compared to the baseline \texttt{Transformer + Visual Encoder} (0.05 FPS).

\begin{wraptable}{l}{0.48\textwidth}
  \centering
  \resizebox{\linewidth}{!}{
  \begin{tabular}{lccc}
  \toprule
  Method & FPS $\uparrow$ & Power (W) $\downarrow$ & WER (\%) $\downarrow$ \\
  \midrule
  Transformer + Visual Encoder & 0.05 & 26.22 & 43.26 \\
  SignX (CSLR in Pixel Space) & 0.57 & 29.20 & 32.78 \\
  SignX (CSLR in Latent Space) & 2.42 & 24.58 & 26.59 \\
  \bottomrule
  \end{tabular}
  }
  \vspace{-2pt}
  \caption{\textbf{Inference Efficiency on ASLLRP:} SignX achieves faster performance by operating in latent space, during the 800-step training process.}
  \label{tab:efficiency}
  \vspace{-24pt}
\end{wraptable}

Beyond throughput, our latent space approach exhibits the highest energy efficiency with a power consumption of \texttt{24.58 W}, whereas the pixel-space model incurs a higher overhead of 29.20 W. Crucially, this gain in efficiency does not come at the cost of accuracy; SignX in latent space yields the lowest WER of \texttt{26.59\%}, marking a significant improvement over the baseline's 43.26\%. These findings validate that operating within a compact, pose-rich latent space effectively mitigates the computational burden of high-dimensional video processing while simultaneously enhancing the robustness of continuous sign language recognition.

\section{Discussion}

SignX achieves a significant performance ceiling by adopting a \textit{structured, modular training paradigm}. While this approach involves a sequence of specialized optimization steps compared to monolithic end-to-end models, this design choice is a \textit{deliberate and beneficial trade-off}. Drawing an analogy to \textit{Latent Diffusion Models}, components like the VAE and UNet can be decoupled or substituted; although they constitute a single system, the modules remain functionally separable. This modularity is a primary driver of their popularity, as it significantly \textit{alleviates the computational and training burden}. Our architecture follows the same principle, ensuring that the \textit{overall net gain} in efficiency and stability outweighs the orchestration complexity.
\section{Conclusion}
\label{sec:conclusion}

This paper presents \textbf{SignX}, a high-efficiency framework for CSLR and SLT. 
\textcolor{purple}{(i)} By leveraging Vision Transformers (ViT) \cite{dosovitskiy2020vit}, we construct a unified and information-dense latent space that successfully integrates heterogeneous pose representations from five state-of-the-art estimators. 
\textcolor{purple}{(ii)} This latent space is further refined through multi-faceted techniques for compression, structural organization, and semantic alignment, establishing a robust sign-specific representation space that effectively maps raw sign language videos to latent embeddings. 
\textcolor{purple}{(iii)} Furthermore, we perform continuous recognition within this compressed, pose-rich space by employing a multi-stage modeling approach that combines ResNet-driven \cite{zhang2023sltunet} temporal feature distillation with Transformer-based sequence refinement, supplemented by latent optimization strategies such as covariance whitening and adaptive pruning. 
\textcolor{purple}{(iv)} SignX significantly enhances recognition accuracy while drastically reducing computational overhead. Extensive experimental validation across four major benchmarks
—demonstrates that SignX achieves SOTA performance across all metrics.
\clearpage
\bibliographystyle{splncs04}
\bibliography{ref/main,ref/sds,ref/meta,ref/llm,ref/rl,ref/slt,ref/how2sign}

@String(IJCV  = {IJCV})

@String(CVPR  = {CVPR})

@String(ICCV  = {ICCV})

@String(ECCV  = {ECCV})

@String(NIPS  = {NeurIPS})

@String(BMVC  =	{BMVC})

@String(TOG   = {ACM TOG})

@String(ICASSP=	{ICASSP})

@String(ICLR  = {ICLR})

@String(AAAI = {AAAI})

@String(IJCV = {Int. J. Comput. Vis.})

@String(CVPR= {IEEE Conf. Comput. Vis. Pattern Recog.})

@String(ICCV= {Int. Conf. Comput. Vis.})

@String(ECCV= {Eur. Conf. Comput. Vis.})

@String(NIPS= {Adv. Neural Inform. Process. Syst.})

@String(BMVC= {Brit. Mach. Vis. Conf.})

@String(TOG= {ACM Trans. Graph.})

@String(ICLR = {Int. Conf. Learn. Represent.})

@book{gloss,
  title={Grammar, gesture, and meaning in American Sign Language},
  author={Liddell, Scott K and others},
  year={2003},
  publisher={Cambridge University Press}
}

@InProceedings{SIGNUM,
  Title = {SIGNUM Database: Video Corpus for Signer-Independent Continuous Sign Language Recognition},
  Author= {Von Agris, U. and Kraiss, K.-F.},
  Booktitle= {Workshop on Representation and Processing of Sign Languages},
  Year= {2010},
  Pages= {243-246}
}

@inproceedings{SLTranslation,
  title={Neural Sign Language Translation},
  author={Cihan Camgoz, Necati and Hadfield, Simon and Koller, Oscar and Ney, Hermann and Bowden, Richard},
  booktitle={CVPR},
  pages={7784--7793},
  year={2018}
}

@misc{gloss-informal,
  title = {Gloss: transcription symbols},
  author = {Jolanta Lapiak},
  howpublished = {\url{https://www.handspeak.com/learn/index.php?id=3}},
  note = {Accessed: 2019-08-20}
}

@inproceedings{WordsAreOurGlosses,
  title={Neural Sign Language Synthesis: Words Are Our Glosses},
  author={Zelinka, Jan and Kanis, Jakub},
  booktitle={The IEEE Winter Conference on Applications of Computer Vision},
  pages={3395--3403},
  year={2020}
}

@inproceedings{signor,
  title={Compiling the Slovene Sign Language Corpus},
  author={Vintar, {\v{S}}pela and Jerko, Bo{\v{s}}tjan and Kulovec, Marjetka},
  booktitle={5th Workshop on the Representation and Processing of Sign Languages: Interactions between Corpus and Lexicon. Language Resources and Evaluation Conference (LREC)},
  volume={5},
  pages={159--162},
  year={2012}
}

@inproceedings{JAsigning,
  title={An open web platform for rule-based speech-to-sign translation},
  author={Rayner, Manny and Bouillon, Pierrette and Ebling, Sarah and Gerlach, Johanna and Strasly, Irene and Tsourakis, Nikos},
  booktitle={54th Annual Meeting of the Association for Computational Linguistics},
  volume={2},
  pages={162--168},
  year={2016}
}

@article{Katz-IEEE-1987-estimation,
  author    = {Slava M. Katz},
  title     = {Estimation of probabilities from sparse data for the language model
               component of a speech recognizer},
  journal   = {{IEEE} Trans. Acoust. Speech Signal Process.},
  volume    = {35},
  number    = {3},
  pages     = {400--401},
  year      = {1987},
}

@article{Zhang-arxiv-2022-Multimodal,
  author    = {Zhuosheng Zhang and
               Aston Zhang and
               Mu Li and
               Hai Zhao and
               George Karypis and
               Alex Smola},
  title     = {Multimodal Chain-of-Thought Reasoning in Language Models},
  journal   = {CoRR},
  volume    = {abs/2302.00923},
  year      = {2023},
  
}

@misc{Chen-arxiv-2023-Robust,
  author = {Chen, Xuanting and Ye, Junjie and Zu, Can and Xu, Nuo and Zheng, Rui and Peng, Minlong and Zhou, Jie and Gui, Tao and Zhang, Qi and Huang, Xuanjing},
  title = {How Robust is GPT-3.5 to Predecessors? A Comprehensive Study on Language Understanding Tasks},
  publisher = {arXiv},
  year = {2023}
}

@article{Lu-2023-arXiv-multimodal,
  author       = {Yujie Lu and
                  Pan Lu and
                  Zhiyu Chen and
                  Wanrong Zhu and
                  Xin Eric Wang and
                  William Yang Wang},
  title        = {Multimodal Procedural Planning via Dual Text-Image Prompting},
  journal      = {CoRR},
  volume       = {abs/2305.01795},
  year         = {2023},
}

@misc{neidle2022asl,
    title={{ASL} {V}ideo {C}orpora \& {S}ign {B}ank: {R}esources {A}vailable through the {A}merican {S}ign {L}anguage {L}inguistic {R}esearch {P}roject ({ASLLRP})},
    author={Carol Neidle and Augustine Opoku and Dimitris Metaxas},
    year={2022},
    eprint={2201.07899},
    archivePrefix={arXiv},
    primaryClass={cs.CL}
}

@INPROCEEDINGS{7780459,
  author={He, Kaiming and Zhang, Xiangyu and Ren, Shaoqing and Sun, Jian},
  booktitle={2016 IEEE Conference on Computer Vision and Pattern Recognition (CVPR)}, 
  title={Deep Residual Learning for Image Recognition}, 
  year={2016},
  volume={},
  number={},
  pages={770-778},
  doi={10.1109/CVPR.2016.90}}

@misc{zhang2025largesignlanguagemodels,
      title={Large Sign Language Models: Toward 3D American Sign Language Translation}, 
      author={Sen Zhang and Xiaoxiao He and Di Liu and Zhaoyang Xia and Mingyu Zhao and Chaowei Tan and Vivian Li and Bo Liu and Dimitris N. Metaxas and Mubbasir Kapadia},
      year={2025},
      eprint={2511.08535},
      archivePrefix={arXiv},
      primaryClass={cs.CV},
      url={https://arxiv.org/abs/2511.08535}, 
}

@article{kingma2013auto,
  title={Auto-encoding variational bayes},
  author={Kingma, Diederik P and Welling, Max},
  journal={arXiv preprint arXiv:1312.6114},
  year={2013}
}

@misc{simeoni2025dinov3,
  title={{DINOv3}},
  author={Sim{\'e}oni, Oriane and Vo, Huy V. and Seitzer, Maximilian and Baldassarre, Federico and Oquab, Maxime and Jose, Cijo and Khalidov, Vasil and Szafraniec, Marc and Yi, Seungeun and Ramamonjisoa, Micha{\"e}l and Massa, Francisco and Haziza, Daniel and Wehrstedt, Luca and Wang, Jianyuan and Darcet, Timoth{\'e}e and Moutakanni, Th{\'e}o and Sentana, Leonel and Roberts, Claire and Vedaldi, Andrea and Tolan, Jamie and Brandt, John and Couprie, Camille and Mairal, Julien and J{\'e}gou, Herv{\'e} and Labatut, Patrick and Bojanowski, Piotr},
  year={2025},
  eprint={2508.10104},
  archivePrefix={arXiv},
  primaryClass={cs.CV},
  url={https://arxiv.org/abs/2508.10104},
}

@InProceedings{10.1007/978-3-319-49409-8_7,
author="Lea, Colin
and Vidal, Ren{\'e}
and Reiter, Austin
and Hager, Gregory D.",
editor="Hua, Gang
and J{\'e}gou, Herv{\'e}",
title="Temporal Convolutional Networks: A Unified Approach to Action Segmentation",
booktitle="Computer Vision -- ECCV 2016 Workshops",
year="2016",
publisher="Springer International Publishing",
address="Cham",
pages="47--54",
isbn="978-3-319-49409-8"
}

@inproceedings{
zhang2023sltunet,
title={{SLTUNET}: A Simple Unified Model for Sign Language Translation},
author={Biao Zhang and Mathias M{\"u}ller and Rico Sennrich},
booktitle={The Eleventh International Conference on Learning Representations },
year={2023},
url={https://openreview.net/forum?id=EBS4C77p_5S}
}

@article{chen2022two,
title={Two-Stream Network for Sign Language Recognition and Translation},
  author={Chen, Yutong and Zuo, Ronglai and Wei, Fangyun and Wu, Yu and Liu, Shujie and Mak, Brian},
  journal={NeurIPS},
  year={2022}
}

@misc{fang2025stablesignerhierarchicalsign,
      title={Stable Signer: Hierarchical Sign Language Generative Model},
      author={Sen Fang and Yalin Feng and Hongbin Zhong and Yanxin Zhang and Dimitris N. Metaxas},
      year={2025},
      eprint={2512.04048},
      archivePrefix={arXiv},
      primaryClass={cs.CV},
      url={https://arxiv.org/abs/2512.04048},
}

@article{li2025uni,
  title={Uni-Sign: Toward Unified Sign Language Understanding at Scale},
  author={Li, Zecheng and Zhou, Wengang and Zhao, Weichao and Wu, Kepeng and Hu, Hezhen and Li, Houqiang},
  journal={arXiv preprint arXiv:2501.15187},
  year={2025}
}

@misc{fang2025streamflowtheoryalgorithmimplementation,
      title={StreamFlow: Theory, Algorithm, and Implementation for High-Efficiency Rectified Flow Generation}, 
      author={Sen Fang and Hongbin Zhong and Yalin Feng and Dimitris N. Metaxas},
      year={2025},
      eprint={2511.22009},
      archivePrefix={arXiv},
      primaryClass={cs.CV},
      url={https://arxiv.org/abs/2511.22009}, 
}

@article{KLAKOW200219,
title = {Testing the correlation of word error rate and perplexity},
journal = {Speech Communication},
volume = {38},
number = {1},
pages = {19-28},
year = {2002},
issn = {0167-6393},
doi = {https://doi.org/10.1016/S0167-6393(01)00041-3},
url = {https://www.sciencedirect.com/science/article/pii/S0167639301000413},
author = {Dietrich Klakow and Jochen Peters}
}

@inproceedings{zuo2022c2slr,
  title={C2slr: Consistency-enhanced Continuous Sign Language Recognition},
  author={Zuo, Ronglai and Mak, Brian},
  booktitle={CVPR},
  pages={5131--5140},
  year={2022}
}

@article{zhou2021spatial,
  title={Spatial-temporal multi-cue network for Sign Language Recognition and Translation},
  author={Zhou, Hao and Zhou, Wengang and Zhou, Yun and Li, Houqiang},
  journal={IEEE Transactions on Multimedia},
  volume={24},
  pages={768--779},
  year={2021},
  publisher={IEEE}
}

@inproceedings{hu2023continuous,
  title={Continuous Sign Language Recognition with Correlation Network},
  author={Hu, Lianyu and Gao, Liqing and Liu, Zekang and Feng, Wei},
  booktitle={Proceedings of the IEEE/CVF International Conference on Computer Vision},
  year={2023},
}

@article{2020t5,
  author  = {Colin Raffel and Noam Shazeer and Adam Roberts and Katherine Lee and Sharan Narang and Michael Matena and Yanqi Zhou and Wei Li and Peter J. Liu},
  title   = {Exploring the Limits of Transfer Learning with a Unified Text-to-Text Transformer},
  journal = {Journal of Machine Learning Research},
  year    = {2020},
  volume  = {21},
  number  = {140},
  pages   = {1-67},
  url     = {http://jmlr.org/papers/v21/20-074.html}
}

@INPROCEEDINGS{710744,
  author={Vogler, C. and Metaxas, D.},
  booktitle={Sixth International Conference on Computer Vision (IEEE Cat. No.98CH36271)}, 
  title={ASL recognition based on a coupling between HMMs and 3D motion analysis}, 
  year={1998},
  volume={},
  number={},
  pages={363-369},
  doi={10.1109/ICCV.1998.710744}}

@INPROCEEDINGS{8578910,
  author={Camgoz, Necati Cihan and Hadfield, Simon and Koller, Oscar and Ney, Hermann and Bowden, Richard},
  booktitle={2018 IEEE/CVF Conference on Computer Vision and Pattern Recognition}, 
  title={Neural Sign Language Translation}, 
  year={2018},
  volume={},
  number={},
  pages={7784-7793},
  doi={10.1109/CVPR.2018.00812}}

@ARTICLE{10542663,
  author={Guo, Leming and Xue, Wanli and Liu, Bo and Zhang, Kaihua and Yuan, Tiantian and Metaxas, Dimitris},
  journal={IEEE Transactions on Image Processing}, 
  title={Gloss Prior Guided Visual Feature Learning for Continuous Sign Language Recognition}, 
  year={2024},
  volume={33},
  number={},
  pages={3486-3495},
  doi={10.1109/TIP.2024.3404869}}

@INPROCEEDINGS{799236,
  author={Metaxas, D.},
  booktitle={Proceedings International Workshop on Recognition, Analysis, and Tracking of Faces and Gestures in Real-Time Systems. In Conjunction with ICCV'99 (Cat. No.PR00378)}, 
  title={Deformable model and HMM-based tracking, analysis and recognition of gestures and faces}, 
  year={1999},
  volume={},
  number={},
  pages={136-140},
  doi={10.1109/RATFG.1999.799236}}

@ARTICLE{9556136,
  author={Tang, Shengeng and Guo, Dan and Hong, Richang and Wang, Meng},
  journal={IEEE Transactions on Multimedia}, 
  title={Graph-Based Multimodal Sequential Embedding for Sign Language Translation}, 
  year={2022},
  volume={24},
  number={},
  pages={4433-4445},
  doi={10.1109/TMM.2021.3117124}}

@inproceedings{neidle2018,
title={{NEW} {Shared} \& {I}nterconnected {ASL R}esources: {S}ign{S}tream® 3 {S}oftware; {DAI} 2 for {W}eb {A}ccess to {L}inguistically {A}nnotated {V}ideo {C}orpora; and a {S}ign {B}ank},
author={Neidle, Carol and Opoku, Augustine and Dimitriadis, Gregory and Metaxas, Dimitris},
booktitle={{L}anguage {R}esources and {E}valuation. 8th {W}orkshop on the {R}epresentation and {P}rocessing of {S}ign {L}anguages: {I}nvolving the {L}anguage {C}ommunity. {M}iyazaki, {J}apan, 2018-05-12},
pages={147-154}
}

@inproceedings{yang2023effective,
  title={Effective whole-body pose estimation with two-stages distillation},
  author={Yang, Zhendong and Zeng, Ailing and Yuan, Chun and Li, Yu},
  booktitle={Proceedings of the IEEE/CVF International Conference on Computer Vision},
  pages={4210--4220},
  year={2023}
}

@misc{yosinski2014transferable,
    title={How transferable are features in deep neural networks?},
    author={Jason Yosinski and Jeff Clune and Yoshua Bengio and Hod Lipson},
    year={2014},
    eprint={1411.1792},
    archivePrefix={arXiv},
    primaryClass={cs.LG}
}

@misc{sutskever2014sequence,
    title={Sequence to Sequence Learning with Neural Networks},
    author={Ilya Sutskever and Oriol Vinyals and Quoc V. Le},
    year={2014},
    eprint={1409.3215},
    archivePrefix={arXiv},
    primaryClass={cs.CL}
}

@misc{rombach2021highresolution,
      title={High-Resolution Image Synthesis with Latent Diffusion Models}, 
      author={Robin Rombach and Andreas Blattmann and Dominik Lorenz and Patrick Esser and Björn Ommer},
      year={2021},
      eprint={2112.10752},
      archivePrefix={arXiv},
      primaryClass={cs.CV}
}

@inproceedings{li2020word,
      title={Word-level Deep Sign Language Recognition from Video: A New Large-scale Dataset and Methods Comparison},
      author={Li, Dongxu and Rodriguez, Cristian and Yu, Xin and Li, Hongdong},
      booktitle={The IEEE Winter Conference on Applications of Computer Vision},
      pages={1459--1469},
      year={2020}
    }

@misc{asllrp2025signbank,
  author = {Neidle, Carol and Metaxas, Dimitris},
  title = {American Sign Language Linguistic Research Project (ASLLRP) Sign Bank},
  howpublished = {\url{https://dai.cs.rutgers.edu/dai/s/signbank}},
  note = {ASLLRP Continuous Signing Corpora, version 3},
  year = {2025},
  month = {June},
  organization = {Boston and Rutgers Universities},
  copyright = {© 2022-2025}
}

@misc{neidle2022aslvideocorpora,
      title={ASL Video Corpora \& Sign Bank: Resources Available through the American Sign Language Linguistic Research Project (ASLLRP)}, 
      author={Carol Neidle and Augustine Opoku and Dimitris Metaxas},
      year={2022},
      eprint={2201.07899},
      archivePrefix={arXiv},
      primaryClass={cs.CL},
      url={https://arxiv.org/abs/2201.07899}, 
}

@inproceedings{zhou-etal-2024-multimodal,
    title = "A Multimodal Spatio-Temporal {GCN} Model with Enhancements for Isolated Sign Recognition",
    author = "Zhou, Yang  and
      Xia, Zhaoyang  and
      Chen, Yuxiao  and
      Neidle, Carol  and
      Metaxas, Dimitris N.",
    editor = "Efthimiou, Eleni  and
      Fotinea, Stavroula-Evita  and
      Hanke, Thomas  and
      Hochgesang, Julie A.  and
      Mesch, Johanna  and
      Schulder, Marc",
    booktitle = "Proceedings of the LREC-COLING 2024 11th Workshop on the Representation and Processing of Sign Languages: Evaluation of Sign Language Resources",
    month = may,
    year = "2024",
    address = "Torino, Italia",
    publisher = "ELRA and ICCL",
    url = "https://aclanthology.org/2024.signlang-1.45/",
    pages = "408--419"
}

@inproceedings{mao2023cross,
  title={Cross-entropy loss functions: Theoretical analysis and applications},
  author={Mao, Anqi and Mohri, Mehryar and Zhong, Yutao},
  booktitle={International conference on Machine learning},
  pages={23803--23828},
  year={2023},
  organization={PMLR}
}

@article{dosovitskiy2020vit,
  title={An Image is Worth 16x16 Words: Transformers for Image Recognition at Scale},
  author={Dosovitskiy, Alexey and Beyer, Lucas and Kolesnikov, Alexander and Weissenborn, Dirk and Zhai, Xiaohua and Unterthiner, Thomas and  Dehghani, Mostafa and Minderer, Matthias and Heigold, Georg and Gelly, Sylvain and Uszkoreit, Jakob and Houlsby, Neil},
  journal={ICLR},
  year={2021}
}

@article{ho2020denoising,
    title={Denoising Diffusion Probabilistic Models},
    author={Jonathan Ho and Ajay Jain and Pieter Abbeel},
    year={2020},
    journal={arXiv preprint arxiv:2006.11239}
}

@article{khirodkar2024sapiens,
  title={Sapiens: Foundation for Human Vision Models},
  author={Khirodkar, Rawal and Bagautdinov, Timur and Martinez, Julieta and Zhaoen, Su and James, Austin and Selednik, Peter and Anderson, Stuart and Saito, Shunsuke},
  journal={arXiv preprint arXiv:2408.12569},
  year={2024}
}

@misc{zavadski2024primedepth,
    title={PrimeDepth: Efficient Monocular Depth Estimation with a Stable Diffusion Preimage}, 
    author={Denis Zavadski and Damjan Kalšan and Carsten Rother},
    year={2024},
    eprint={2409.09144},
    archivePrefix={arXiv},
    primaryClass={cs.CV},
    url={https://arxiv.org/abs/2409.09144}, 
}

@inproceedings{yin2022mlslt,
  title={MLSLT: Towards Multilingual Sign Language Translation},
  author={Yin, Aoxiong and Zhao, Zhou and Jin, Weike and Zhang, Meng and Zeng, Xingshan and He, Xiaofei},
  booktitle={Proceedings of the IEEE/CVF Conference on Computer Vision and Pattern Recognition},
  pages={5109--5119},
  year={2022}
}

@article{neidle2022alternative,
  title={Why alternative gloss labels will increase the value of the WLASL dataset},
  author={Neidle, Carol and Ballard, Carey},
  year={2022},
  publisher={Boston University American Sign Language Linguistic Research Project}
}

@misc{hu2024corrnetsignlanguagerecognition,
      title={CorrNet+: Sign Language Recognition and Translation via Spatial-Temporal Correlation}, 
      author={Lianyu Hu and Wei Feng and Liqing Gao and Zekang Liu and Liang Wan},
      year={2024},
      eprint={2404.11111},
      archivePrefix={arXiv},
      primaryClass={cs.CV},
      url={https://arxiv.org/abs/2404.11111}, 
}

@misc{fang2025signllmsignlanguageproduction,
      title={SignLLM: Sign Language Production Large Language Models}, 
      author={Sen Fang and Chen Chen and Lei Wang and Ce Zheng and Chunyu Sui and Yapeng Tian},
      year={2025},
      eprint={2405.10718},
      archivePrefix={arXiv},
      primaryClass={cs.CV},
      url={https://arxiv.org/abs/2405.10718}, 
}

@misc{fang2025signdiffdiffusionmodelamerican,
      title={SignDiff: Diffusion Model for American Sign Language Production}, 
      author={Sen Fang and Chunyu Sui and Yanghao Zhou and Xuedong Zhang and Hongbin Zhong and Yapeng Tian and Chen Chen},
      year={2025},
      eprint={2308.16082},
      archivePrefix={arXiv},
      primaryClass={cs.CV},
      url={https://arxiv.org/abs/2308.16082}, 
}

@InProceedings{slt-how2sign-wicv2023,
author = {Laia Tarrés and Gerard I. Gállego and Amanda Duarte and Jordi Torres 
          and Xavier Giró-i-Nieto},
title = {Sign Language Translation from Instructional Videos},
booktitle = {Proceedings of the IEEE/CVF Conference on Computer Vision and Pattern Recognition 
            (CVPR) Workshops},
year = {2023}
}

@INPROCEEDINGS{7298594,
  author={Szegedy, Christian and Wei Liu and Yangqing Jia and Sermanet, Pierre and Reed, Scott and Anguelov, Dragomir and Erhan, Dumitru and Vanhoucke, Vincent and Rabinovich, Andrew},
  booktitle={2015 IEEE Conference on Computer Vision and Pattern Recognition (CVPR)}, 
  title={Going deeper with convolutions}, 
  year={2015},
  volume={},
  number={},
  pages={1-9},
  doi={10.1109/CVPR.2015.7298594}}

@Article {koller15:cslr,
author= {Koller, Oscar and Forster, Jens and Ney, Hermann},
title= {Continuous Sign Language Recognition: Towards large vocabulary statistical recognition systems handling multiple signers},
journal= {Computer Vision and Image Understanding},
pages= {108-125},
volume= {141},
year= 2015,
month= dec,
journallink= {http://www.journals.elsevier.com/computer-vision-and-image-understanding/}
}

@inproceedings{guler2018densepose,
  title={Densepose: Dense human pose estimation in the wild},
  author={ G{\"u}ler, R{\i}za Alp and Neverova, Natalia and Kokkinos, Iasonas},
  booktitle={Proceedings of the IEEE Conference on Computer Vision and Pattern Recognition},
  pages={7297--7306},
  year={2018}
}

@InProceedings{Bohacek_2022_WACV,
    author    = {Boh\'a\v{c}ek, Maty\'a\v{s} and Hr\'uz, Marek},
    title     = {Sign Pose-Based Transformer for Word-Level Sign Language Recognition},
    booktitle = {Proceedings of the IEEE/CVF Winter Conference on Applications of Computer Vision (WACV) Workshops},
    month     = {January},
    year      = {2022},
    pages     = {182-191}
}

@misc{cai2023smplerx,
      title={SMPLer-X: Scaling Up Expressive Human Pose and Shape Estimation}, 
      author={Zhongang Cai and Wanqi Yin and Ailing Zeng and Chen Wei and Qingping Sun and Yanjun Wang and Hui En Pang and Haiyi Mei and Mingyuan Zhang and Lei Zhang and Chen Change Loy and Lei Yang and Ziwei Liu},
      year={2023},
      eprint={2309.17448},
      archivePrefix={arXiv},
      primaryClass={cs.CV}
}

@inproceedings{smpl,
    title       =   {{SMPL}: {A} skinned multi-person linear model},
    author      =   {Loper, Matthew and Mahmood, Naureen and Romero, Javier and Pons-Moll, Gerard and Black, Michael J.},
    booktitle   =   TOG,
    year        =   {2015}
}

@inproceedings{zuo2023natural,
  title={Natural language-assisted Sign Language Recognition},
  author={Zuo, Ronglai and Wei, Fangyun and Mak, Brian},
  booktitle={Proceedings of the IEEE/CVF conference on computer vision and pattern recognition},
  pages={14890--14900},
  year={2023}
}

@article{zhou2024scaling,
  title={Scaling up multimodal pre-training for Sign Language Understanding},
  author={Zhou, Wengang and Zhao, Weichao and Hu, Hezhen and Li, Zecheng and Li, Houqiang},
  journal={IEEE Transactions on Pattern Analysis and Machine Intelligence},
  year={2025},
  publisher={IEEE}
}

@misc{pu2025sigmasemanticallyinformativepretraining,
      title={Sigma: Semantically Informative Pre-training for Skeleton-based Sign Language Understanding}, 
      author={Muxin Pu and Mei Kuan Lim and Chun Yong Chong and Chen Change Loy},
      year={2025},
      eprint={2509.21223},
      archivePrefix={arXiv},
      primaryClass={cs.CV},
      url={https://arxiv.org/abs/2509.21223}, 
}

@inproceedings{hu2021signbert,
  title={SignBERT: Pre-training of hand-model-aware representation for sign language recognition},
  author={Hu, Hezhen and Zhao, Weichao and Zhou, Wengang and Wang, Yuechen and Li, Houqiang},
  booktitle={Proceedings of the IEEE/CVF international conference on computer vision},
  pages={11087--11096},
  year={2021}
}

@inproceedings{zhao2023best,
  title={BEST: BERT pre-training for sign language recognition with coupling tokenization},
  author={Zhao, Weichao and Hu, Hezhen and Zhou, Wengang and Shi, Jiaxin and Li, Houqiang},
  booktitle={Proceedings of the AAAI conference on artificial intelligence},
  volume={37},
  number={3},
  pages={3597--3605},
  year={2023}
}

@article{hu2023signbert+,
  title={SignBERT+: Hand-model-aware self-supervised pre-training for sign language understanding},
  author={Hu, Hezhen and Zhao, Weichao and Zhou, Wengang and Li, Houqiang},
  journal={IEEE Transactions on Pattern Analysis and Machine Intelligence},
  volume={45},
  number={9},
  pages={11221--11239},
  year={2023},
  publisher={IEEE}
}

@inproceedings{zhang2020fusionnet,
	title={Deep FusionNet for Point Cloud Semantic Segmentation},
	author={Zhang, Feihu and Fang, Jin and Wah, Benjamin and Torr, Philip},
	booktitle=ECCV,
	year = {2020}
}

@article{chen2020simple,
	title={A Simple Framework for Contrastive Learning of Visual Representations},
	author={Chen, Ting and Kornblith, Simon and Norouzi, Mohammad and Hinton, Geoffrey},
	journal={arXiv preprint arXiv:2002.05709},
	year={2020}
}

@inproceedings{qiu2021dense,
	title={Dense-resolution network for point cloud classification and segmentation},
	author={Qiu, Shi and Anwar, Saeed and Barnes, Nick},
	booktitle={WACV},
	pages={3813--3822},
	year={2021}
}

@inproceedings{chiang2019unified,
	title     = {A unified point-based framework for 3d segmentation},
	author    = {Chiang, Hung-Yueh and Lin, Yen-Liang and Liu, Yueh-Cheng and Hsu, Winston H},
	booktitle = {3DV},
	year      = {2019}
}

@article{dong2022act,
	title={Autoencoders as Cross-Modal Teachers: Can Pretrained 2D Image Transformers Help 3D Representation Learning?},
	author={Dong, Runpei and Qi, Zekun and Zhang, Linfeng and Zhang, Junbo and Sun, Jianjian and Ge, Zheng and Yi, Li and Ma, Kaisheng},
	journal={arXiv preprint arXiv:2212.08320},
	year={2022}
}

@inproceedings{xie2021simmim,
	title={SimMIM: A Simple Framework for Masked Image Modeling},
	author={Xie, Zhenda and Zhang, Zheng and Cao, Yue and Lin, Yutong and Bao, Jianmin and Yao, Zhuliang and Dai, Qi and Hu, Han},
	booktitle={CVPR},
	year={2022}
}

@inproceedings{he2022masked,
	title={Masked autoencoders are scalable vision learners},
	author={He, Kaiming and Chen, Xinlei and Xie, Saining and Li, Yanghao and Doll{\'a}r, Piotr and Girshick, Ross},
	booktitle={Proceedings of the IEEE/CVF Conference on Computer Vision and Pattern Recognition},
	pages={16000--16009},
	year={2022}
}

@article{xie2021segformer,
	title={SegFormer: Simple and efficient design for semantic segmentation with transformers},
	author={Xie, Enze and Wang, Wenhai and Yu, Zhiding and Anandkumar, Anima and Alvarez, Jose M and Luo, Ping},
	journal={Advances in Neural Information Processing Systems},
	volume={34},
	pages={12077--12090},
	year={2021}
}

@article{liu2022bevfusion,
	title={BEVFusion: Multi-Task Multi-Sensor Fusion with Unified Bird's-Eye View Representation},
	author={Liu, Zhijian and Tang, Haotian and Amini, Alexander and Yang, Xinyu and Mao, Huizi and Rus, Daniela and Han, Song},
	journal={arXiv preprint arXiv:2205.13542},
	year={2022}
}

@inproceedings{hershey2017cnn,
	title={CNN architectures for large-scale audio classification},
	author={Hershey, Shawn and Chaudhuri, Sourish and Ellis, Daniel PW and Gemmeke, Jort F and Jansen, Aren and Moore, R Channing and Plakal, Manoj and Platt, Devin and Saurous, Rif A and Seybold, Bryan and others},
	booktitle={2017 ieee international conference on acoustics, speech and signal processing (icassp)},
	pages={131--135},
	year={2017},
	organization={IEEE}
}

@inproceedings{lu_unified-io_2023,
	title = {{UNIFIED}-{IO}: {A} {Unified} {Model} for {Vision}, {Language}, and {Multi}-modal {Tasks}},
	shorttitle = {{UNIFIED}-{IO}},
	url = {https://openreview.net/forum?id=E01k9048soZ},
	language = {en},
	urldate = {2023-04-27},
	author = {Lu, Jiasen and Clark, Christopher and Zellers, Rowan and Mottaghi, Roozbeh and Kembhavi, Aniruddha},
	month = feb,
	year = {2023},
}

@inproceedings{saunders2020progressive,
	title		=	{{Progressive Transformers for End-to-End Sign Language Production}},
	author		=	{Saunders, Ben and Camg{\"o}z, Necati Cihan and Bowden, Richard},
	booktitle   =   {Proceedings of the European Conference on Computer Vision (ECCV)},
	year		=	{2020}}

@article{saunders2021continuous,
	title		=	{{Continuous 3D Multi-Channel Sign Language Production via Progressive Transformers and Mixture Density Networks}},
	author		=	{Saunders, Ben and Camg{\"o}z, Necati Cihan and Bowden, Richard},
	journal     =   {International Journal of Computer Vision (IJCV)},
	year		=	{2021}}

@inproceedings{saunders2021mixed,
	title		=	{{Mixed SIGNals: Sign Language Production via a Mixture of Motion Primitives}},
	author		=	{Saunders, Ben and Camg{\"o}z, Necati Cihan and Bowden, Richard},
	booktitle   =   {Proceedings of the International Conference on Computer Vision (ICCV)},
	year		=	{2021}}

@article{saunders2021skeletal,
    title       =   {{Skeletal Graph Self-Attention: Embedding a Skeleton Inductive Bias into Sign Language Production}},
    author      =   {Saunders, Ben and Camgoz, Necati Cihan and Bowden, Richard},
    journal     =   {arXiv preprint arXiv:2112.05277},
    year        =   {2021}}

@article{stoll2020text2sign,
    title       =   {{Text2Sign: Towards Sign Language Production using Neural Machine Translation and Generative Adversarial Networks}},
    author      =   {Stoll, Stephanie and Camg{\"o}z, Necati Cihan and Hadfield, Simon and Bowden, Richard},
    journal     =   {International Journal of Computer Vision (IJCV)},
    year        =   {2020}}

@inproceedings{zelinka2020neural,
    title       =   {{Neural Sign Language Synthesis: Words Are Our Glosses}},
    author      =   {Zelinka, Jan and Kanis, Jakub},
    booktitle   =   {The IEEE Winter Conference on Applications of Computer Vision (WACV)},
    year        =   {2020}}

@inproceedings{huang2021towards,
    title       =   {{Towards Fast and High-Quality Sign Language Production}},
    author      =   {Huang, Wencan and Pan, Wenwen and Zhao, Zhou and Tian, Qi},
    booktitle   =   {Proceedings of the 29th ACM International Conference on Multimedia},
    year        =   {2021}}

@inproceedings{cui2017recurrent,
	title		=	{{Recurrent Convolutional Neural Networks for Continuous Sign Language Recognition by Staged Optimization}},
	author		=	{Cui, Runpeng and Liu, Hu and Zhang, Changshui},
	booktitle	=	{Proceedings of the IEEE Conference on Computer Vision and Pattern Recognition (CVPR)},
	year		=	{2017}}

@inproceedings{grobel1997isolated,
    title       =   {{Isolated Sign Language Recognition using Hidden Markov Models}},
    author      =   {Grobel, Kirsti and Assan, Marcell},
    booktitle   =   {IEEE International Conference on Systems, Man, and Cybernetics},
    year        =   {1997}}

@article{koller2020quantitative,
    title       =   {{Quantitative Survey of the State of the Art in Sign Language Recognition}},
    author      =   {Koller, Oscar},
    journal     =   {arXiv preprint arXiv:2008.09918},
    year        =   {2020}}

@inproceedings{kadir2004minimal,
    title       =   {{Minimal Training, Large Lexicon, Unconstrained Sign Language Recognition}},
    author      =   {Kadir, Timor and Bowden, Richard and Ong, Eng-Jon and Zisserman, Andrew},
	booktitle	=	{Proceedings of the British Machine Vision Conference (BMVC)},
    year        =   {2004}}

@inproceedings{cooper2007large,
    title       =   {{Large Lexicon Detection of Sign Language}},
    author      =   {Cooper, Helen and Bowden, Richard},
    booktitle   =   {International Workshop on Human-Computer Interaction},
    year        =   {2007}}

@article{koller2015continuous,
	title		=	{{Continuous Sign Language Recognition: Towards Large Vocabulary Statistical Recognition Systems Handling Multiple Signers}},
	author		=	{Koller, Oscar and Forster, Jens and Ney, Hermann},
	journal		=	{Computer Vision and Image Understanding (CVIU)},
	year		=	{2015}}

@article{ko2019neural,
    title       =   {{Neural Sign Language Translation based on Human Keypoint Estimation}},
    author      =   {Ko, Sang-Ki and Kim, Chang Jo and Jung, Hyedong and Cho, Choongsang},
    journal     =   {Applied Sciences},
    year        =   {2019}}

@inproceedings{camgoz2020sign,
  title         =   {{Sign Language Transformers: Joint End-to-end Sign Language Recognition and Translation}},
  author        =   {Cihan Camg{\"o}z, Necati and Koller, Oscar and Hadfield, Simon and Bowden, Richard},
  booktitle	=	{Proceedings of the IEEE Conference on Computer Vision and Pattern Recognition (CVPR)},
  year          =   {2020}}

@inproceedings{forster2012rwth,
	title		=	{{RWTH-PHOENIX-Weather: A Large Vocabulary Sign Language Recognition and Translation Corpus}},
	author		=	{Forster, Jens and Schmidt, Christoph and Hoyoux, Thomas and Koller, Oscar and Zelle, Uwe and Piater, Justus H and Ney, Hermann},
	booktitle	=	{Proceedings of the International Conference on Language Resources and Evaluation (LREC)},
	year		=	{2012}}

@inproceedings{camgoz2018neural,
	title		=	{{Neural Sign Language Translation}},
	author		=	{Camg{\"o}z, Necati Cihan and Hadfield, Simon and Koller, Oscar and Ney, Hermann and Bowden, Richard},
	booktitle	=	{Proceedings of the IEEE Conference on Computer Vision and Pattern Recognition (CVPR)},
	year		=	{2018}}

@inproceedings{duarte2021how2sign,
    title       =   {{How2Sign: A Large-Scale Multimodal Dataset for Continuous American Sign Language}},
    author      =   {Duarte, Amanda and Palaskar, Shruti and Ventura, Lucas and Ghadiyaram, Deepti and DeHaan, Kenneth and Metze, Florian and Torres, Jordi and Giro-i-Nieto, Xavier},
    booktitle   =   {Proceedings of the IEEE/CVF Conference on Computer Vision and Pattern Recognition (CVPR)},
    year        =   {2021}}

@inproceedings{vaswani2017attention,
    title       =   {{Attention Is All You Need}},
    author      =   {Vaswani, Ashish and Shazeer, Noam and Parmar, Niki and Uszkoreit, Jakob and Jones, Llion and Gomez, Aidan N and Kaiser, {\L}ukasz and Polosukhin, Illia},
    booktitle   =   {Advances in Neural Information Processing Systems (NIPS)},
    year        =   {2017}}

@inproceedings{cao2018openpose,
    title       =   {{OpenPose: Realtime Multi-Person 2D Pose Estimation using Part Affinity Fields}},
    author      =   {Zhe Cao and Gines Hidalgo and Tomas Simon and Shih-En Wei and Yaser Sheikh},
    booktitle   =   {Proceedings of the IEEE Conference on Computer Vision and Pattern Recognition (CVPR)},
    year        =   {2017}}

@article{charles2014automatic,
    title       =   {{Automatic and Efficient Human Pose Estimation for Sign Language Videos}},
    author      =   {Charles, James and Pfister, Tomas and Everingham, Mark and Zisserman, Andrew},
    journal     =   {International Journal of Computer Vision (IJCV)},
    year        =   {2014},
    publisher   =   {Springer}}

@inproceedings{wang2018high,
    title       =   {{High-Resolution Image Synthesis and Semantic Manipulation with Conditional GANs}},
    author      =   {Wang, Ting-Chun and Liu, Ming-Yu and Zhu, Jun-Yan and Tao, Andrew and Kautz, Jan and Catanzaro, Bryan},
    booktitle   =   {Proceedings of the IEEE Conference on Computer Vision and Pattern Recognition (CVPR)},
    year        =   {2018}}

@inproceedings{wang2018video,
    title       =   {{Video-to-Video Synthesis}},
    author      =   {Wang, Ting-Chun and Liu, Ming-Yu and Zhu, Jun-Yan and Liu, Guilin and Tao, Andrew and Kautz, Jan and Catanzaro, Bryan},
    booktitle   =   {Advances in Neural Information Processing Systems (NIPS)},
    year        =   {2018}}

@inproceedings{chan2019everybody,
    title       =   {{Everybody Dance Now}},
    author      =   {Chan, Caroline and Ginosar, Shiry and Zhou, Tinghui and Efros, Alexei A},
    booktitle   =   {Proceedings of the IEEE International Conference on Computer Vision (CVPR)},
    year        =   {2019}}

@INPROCEEDINGS{2018_Neural_Sign_Language_Translation,
  author={Camgoz, Necati Cihan and Hadfield, Simon and Koller, Oscar and Ney, Hermann and Bowden, Richard},
  booktitle={2018 IEEE/CVF Conference on Computer Vision and Pattern Recognition}, 
  title={Neural Sign Language Translation}, 
  year={2018},
  pages={7784-7793},
  doi={10.1109/CVPR.2018.00812}}

@inproceedings{zhou2021improving,
  title={Improving sign language translation with monolingual data by sign back-translation},
  author={Zhou, Hao and Zhou, Wengang and Qi, Weizhen and Pu, Junfu and Li, Houqiang},
  booktitle={Proceedings of the IEEE/CVF Conference on Computer Vision and Pattern Recognition},
  pages={1316--1325},
  year={2021}
}

@article{yin-etal-2021-including-signed-languages,
  title={Including signed languages in natural language processing},
  author={Yin, Kayo and Moryossef, Amit and Hochgesang, Julie and Goldberg, Yoav and Alikhani, Malihe},
  journal={Proceedings of the 59th Annual Meeting of the Association for Computational Linguistics and the 11th International Joint Conference on Natural Language Processing},
  year={2021}
}

@article{Zhou2021ImprovingSLT-with-monolingual-CSLDaily,
  title={Improving Sign Language Translation with Monolingual Data by Sign Back-Translation},
  author={Hao Zhou and Wen-gang Zhou and Weizhen Qi and Junfu Pu and Houqiang Li},
  journal={Proceedings of the IEEE/CVF Conference on Computer Vision and Pattern Recognition (CVPR)},
  year={2021},
  pages={1316-1325}
}

@article{Ko-2019-SLT-based-human-keypoint-estimation,
  title={Neural sign language translation based on human keypoint estimation},
  author={Ko, Sang-Ki and Kim, Chang Jo and Jung, Hyedong and Cho, Choongsang},
  journal={Applied sciences},
  volume={9},
  number={13},
  pages={2683},
  year={2019},
  publisher={MDPI}
}

@article{Kim-2022-Keypointbased-SLT-without-glosses,
  title={Keypoint based sign language translation without glosses},
  author={Kim, Youngmin and Kwak, Minji and Lee, Dain and Kim, Yeongeun and Baek, Hyeongboo},
  journal={arXiv preprint arXiv:2204.10511},
  year={2022}
}

@article{shi-etal-2022-openASL,
  title={Open-domain sign language translation learned from online video},
  author={Shi, Bowen and Brentari, Diane and Shakhnarovich, Greg and Livescu, Karen},
  journal={Proceedings of the 2022 Conference on Empirical Methods in Natural Language Processing},
  year={2022}
}

@inproceedings{Duarte_2021_how2sign,
    title={{How2Sign: A Large-scale Multimodal Dataset for Continuous American Sign Language}},
    author={Duarte, Amanda and Palaskar, Shruti and Ventura, Lucas and Ghadiyaram, Deepti and DeHaan, Kenneth and
                   Metze, Florian and Torres, Jordi and Giro-i-Nieto, Xavier},
    booktitle={Proceedings of the IEEE/CVF Conference on Computer Vision and Pattern Recognition (CVPR)},
    year={2021}
}

@inproceedings{bleu,
    title = "{B}leu: a Method for Automatic Evaluation of Machine Translation",
    author = "Papineni, Kishore  and
      Roukos, Salim  and
      Ward, Todd  and
      Zhu, Wei-Jing",
    booktitle = "Proceedings of the 40th Annual Meeting of the Association for Computational Linguistics",
    month = jul,
    year = "2002",
    address = "Philadelphia, Pennsylvania, USA",
    publisher = "Association for Computational Linguistics",
    url = "https://aclanthology.org/P02-1040",
    doi = "10.3115/1073083.1073135",
    pages = "311--318",
}

@article{seq2seq-RNN,
  title={Sequence to sequence learning with neural networks},
  author={Sutskever, Ilya and Vinyals, Oriol and Le, Quoc V},
  journal={Advances in neural information processing systems},
  volume={27},
  year={2014}
}

@inproceedings{cho2021unifying,
  title={Unifying vision-and-language tasks via text generation},
  author={Cho, Jaemin and Lei, Jie and Tan, Hao and Bansal, Mohit},
  booktitle={International Conference on Machine Learning},
  pages={1931--1942},
  year={2021},
  organization={PMLR}
}

@INPROCEEDINGS {carreira-2017-i3d,
author = {J. Carreira and A. Zisserman},
booktitle = {Proceedings of the IEEE/CVF Conference on Computer Vision and Pattern Recognition (CVPR)},
title = {Quo Vadis, Action Recognition? A New Model and the Kinetics Dataset},
year = {2017}
}

@article{MediaPipe,
  title={Mediapipe: A framework for building perception pipelines},
  author={Lugaresi, Camillo and Tang, Jiuqiang and Nash, Hadon and McClanahan, Chris and Uboweja, Esha and Hays, Michael and Zhang, Fan and Chang, Chuo-Ling and Yong, Ming Guang and Lee, Juhyun and others},
  journal={arXiv preprint arXiv:1906.08172},
  year={2019}
}

@misc{pose_format_helper,
    title={Complete Toolkit for working with poses},
    author={Moryossef, Amit},
    howpublished={\url{https://github.com/AmitMY/pose-format/}},
    year={2022}
}

@inproceedings{WLASL,
  title={Word-level deep sign language recognition from video: A new large-scale dataset and methods comparison},
  author={Li, Dongxu and Rodriguez, Cristian and Yu, Xin and Li, Hongdong},
  booktitle={Proceedings of the IEEE/CVF winter conference on applications of computer vision},
  pages={1459--1469},
  year={2020}
}

@inproceedings{chen2022simple,
  title={A simple multi-modality transfer learning baseline for sign language translation},
  author={Chen, Yutong and Wei, Fangyun and Sun, Xiao and Wu, Zhirong and Lin, Stephen},
  booktitle={Proceedings of the IEEE/CVF Conference on Computer Vision and Pattern Recognition},
  pages={5120--5130},
  year={2022}
}

@INPROCEEDINGS{SLT,
  author={Cihan Camgöz, Necati and Koller, Oscar and Hadfield, Simon and Bowden, Richard},
  booktitle={2020 IEEE/CVF Conference on Computer Vision and Pattern Recognition (CVPR)}, 
  title={Sign Language Transformers: Joint End-to-End Sign Language Recognition and Translation}, 
  year={2020}
  }

@article{SLT_EvSignNow,
  title={Everybody sign now: Translating spoken language to photo realistic sign language video},
  author={Saunders, Ben and Camgoz, Necati Cihan and Bowden, Richard},
  journal={arXiv preprint arXiv:2011.09846},
  year={2020}
}

@inproceedings{SLT_pose_amit,
title	= {Real-Time Sign Language Detection using Human Pose Estimation},
author	= {Amit Moryossef and Ioannis Tsochantaridis and Roee Yosef Aharoni and Sarah Ebling and Srini Narayanan},
year	= {2020},
note	= {https://www.slrtp.com/}
}

\appendix
\clearpage

\startcontents



\hypersetup{
     linkcolor=black
}

\printcontents{}{1}{}

\begin{table*}[!h]
\centering
\resizebox{0.99\textwidth}{!}{
\begin{tabular}{lccccccccl}
\toprule
Dataset                                                                               & \multicolumn{3}{c}{Duration(h)} & \multicolumn{3}{c}{Vocabulary(k)}                                              & Annotation Type               & Year  & Domain                 \\
 & train     & val      & test     & \multicolumn{1}{l}{train} & \multicolumn{1}{l}{val} & \multicolumn{1}{l}{test} &                           &       &                        \\ \midrule
KETI~\cite{Ko-2019-SLT-based-human-keypoint-estimation}                              & 20.05     & 2.24     & 5.70     & $\leftarrow$ & 0.49 & $\rightarrow$                                                       & Spoken Text               & 2019  & Emergency situations   \\
PHOENIX-2014T~\cite{2018_Neural_Sign_Language_Translation}                           & 9.2       & 0.6      & 0.7      & 2                         & 0.9                     & 1                        & Spoken Text, Gloss               & 2018  & Weather Forecast       \\
CSL Daily~\cite{Zhou2021ImprovingSLT-with-monolingual-CSLDaily}                      & 20.62     & 1.24     & 1.41     & 2                         & 1.3                     & 1.3                      & Spoken Text, Gloss               & 2021  & Daily life             \\
OpenASL~\cite{shi-etal-2022-openASL}                            & $\leftarrow$ & 288  & $\rightarrow$          & $\leftarrow$ & 33 &$\rightarrow$                                                         & Spoken Text               & 2022  & Youtube (news + vlogs) \\ 
How2Sign~\cite{Duarte_2021_how2sign}                          & 69.6      & 3.9      & 5.6      & 15.6                      & 3.2                     & 3.6                      & Spoken Text               & 2021  & Instructional          \\ 
\midrule
ASLLRP~\cite{neidle2022asl}                          & $\leftarrow$ & 80  & $\rightarrow$          & $\leftarrow$ & 2.1 &$\rightarrow$                                                         & Gloss                 & 2022  & Comprehensive              \\ 
\bottomrule
\end{tabular}
}
\vspace{4pt}
\caption{\textbf{
Comparison of sign language datasets by annotation type and task focus:
} The top five rows primarily focus on translation to spoken word text. However, our main focus is on recognition of ASL signs (identified by ID Gloss labels) from poses and sign language videos. $\leftarrow$ $\rightarrow$ indicate that in some cases only statistics on the whole dataset are provided.}
\label{tab:SLT_datasets}
\vspace{-30pt}
\end{table*}

\section{Background Information}
\label{sec:related}

\subsection{Sign Language Recognition}
Sign language recognition is an important research direction,  at the intersection of computer vision and natural language processing. Early works mainly relied on hand-crafted features and rules to recognize small sign language vocabularies \cite{grobel1997isolated, kadir2004minimal}. Subsequently, methods based on the use of machine learning (HMMs) and 3D human pose and motion analytics were used to recognize ASL \cite{710744,799236}. With the development of deep learning, neural network-based methods began to dominate this field. The earliest deep learning methods used CNN \cite{hershey2017cnn} to extract spatial features from video frames and then used RNN \cite{seq2seq-RNN} to model temporal relationships \cite{cui2017recurrent, koller2015continuous}. Although these methods were simple and direct, they struggled to capture fine-grained dynamic movements \cite{fang2025stablesignerhierarchicalsign}.

Subsequently, researchers began to introduce human pose estimation technology \cite{Katz-IEEE-1987-estimation,Ko-2019-SLT-based-human-keypoint-estimation,charles2014automatic,SLT,SLT_pose_amit,yin2022mlslt}, converting sign language videos into skeleton sequences for recognition, which significantly improved the models' ability to perceive action details \cite{saunders2020progressive, saunders2021continuous}. With the advancement of computer vision technology, various powerful pose estimation methods \cite{cai2023smplerx,yang2023effective} have emerged. However, because of their different data formats and representation methods, finding a way to effectively integrate this heterogeneous information has become an urgent challenge \cite{stoll2020text2sign, huang2021towards,fang2025signllmsignlanguageproduction,fang2025signdiffdiffusionmodelamerican}. As shown in Table \ref{tab:SLT_datasets}, different datasets focus on different output goals. Understanding also requires taking account of the complex syntax of these languages. This involves not only linear sign order, but also interactions with non-manual signals, which play an essential grammatical role in signed languages. Gloss-based representations like ``I-GO-STORE'' can help, in part, to model the unique linguistic rules of a given signed language \cite{camgoz2018neural, camgoz2020sign,gloss,WordsAreOurGlosses,zelinka2020neural,neidle2022alternative,zhou-etal-2024-multimodal,neidle2022aslvideocorpora,zhang2025largesignlanguagemodels}.

\subsection{Pose Estimation}
Pose estimation technology has made significant progress in recent years. SMPLer-X \cite{cai2023smplerx} achieves precise modeling of the body, hands, and face by providing complete 3D human body model parameter estimation \cite{smpl, cao2018openpose}. DWPose \cite{yang2023effective} focuses on 2D keypoint detection, achieving real-time performance while ensuring stability \cite{wang2018video, MediaPipe}. Google's MediaPipe \cite{MediaPipe} provides lightweight but accurate 3D pose prediction, making real-time applications possible \cite{MediaPipe, chan2019everybody}.

Beyond traditional pose estimation, there are also emerging technologies. PrimeDepth \cite{zavadski2024primedepth}, specifically designed for depth estimation, can provide 3D structural information about scenes \cite{qiu2021dense, carreira-2017-i3d}. Sapiens segmentation \cite{khirodkar2024sapiens} provides a new perspective for interpreting signers' fine-grained actions through precise human body part segmentation \cite{guler2018densepose, saunders2021mixed}. These advances provide rich feature representation methods for SL recognition, offering good extensions to the information dimensions of pose estimation \cite{7298594, koller2020quantitative}.

\subsection{Diffusion Models}
Diffusion models have achieved breakthroughs in the generative arena in recent years. From the initial DDPM \cite{ho2020denoising} to Stable Diffusion \cite{rombach2021highresolution}, diffusion models have demonstrated powerful image generation capabilities \cite{saunders2021skeletal, wang2018high}. Researchers have also begun applying diffusion to discrete text sequence generation, with promising results \cite{ko2019neural, Bohacek_2022_WACV}. Particularly in cross-modal conversion tasks, diffusion models have shown excellent performance, providing important references for converting SL videos to text \cite{saunders2020progressive, cooper2007large,zhou-etal-2024-multimodal}.

The success of the stable diffusion model lies in the fact that it conducts learning not at the pixel level but in the latent space \cite{fang2025streamflowtheoryalgorithmimplementation}. This enables its learning to be not only smooth but also with a reduced computational burden. Inspired by its success, our work also applies sign language recognition in the sign latent space, and as a result, we have achieved very good results.

\section{Methodology for Building Sign Language Latent Space}

\paragraph{Dataset Preparation.} Our process of constructing the rich pose latent space utilize a linguistically annotated ASL dataset of continuous signing: the ASLLRP SignStream® 3 Corpus \cite{neidle2022asl}. This dataset contains over 80 hours of American Sign Language (ASL) videos with synchronized front view, side view, and facial close-up recordings. These videos have been linguistically annotated using SignStream® \cite{neidle2018} software 
, providing detailed manually annotated gloss labels, sign types, handshape information, and non-manual grammatical markers. The dataset comprises 2,127 utterances with 17,522 sign tokens, plus an additional set of 542 sentences from DawnSignPress. 

For video processing, we apply five pose extraction models: SMPLer-X \cite{cai2023smplerx} for accurate 3D human body parameters, DWPose \cite{yang2023effective} for real-time 2D keypoint detection, MediaPipe \cite{MediaPipe} for lightweight 3D pose prediction, PrimeDepth \cite{zavadski2024primedepth} for hand depth information, and Sapiens segmentation \cite{khirodkar2024sapiens} for fine-grained body part segmentation. Unlike previous approaches, our preprocessing strategy retains some incomplete or slightly occluded video samples to enhance model robustness to quality variations in real-world scenarios. Videos are standardized to 30fps and processed at uniform spatial resolution. We extract multimodal pose features from all videos to form a unified representation space. 

Sign recognition presents significant challenges in multimodal representation learning, particularly in converting continuous visual gestures into discrete textual descriptions \cite{camgoz2020sign,SLTranslation,SLT_EvSignNow,Kim-2022-Keypointbased-SLT-without-glosses}. Regarding the construction of the posture latent space, We propose a 2-stage training approach based on ViT that systematically learns Pose2Gloss and Video2Pose stages.
\subsection{Stage 1: ViT Pose2Gloss Part Training}

\paragraph{Text Generation Loss.}
Following seq2seq principles \cite{sutskever2014sequence}, we employ a cross-entropy text generation loss \cite{mao2023cross} that measures the probabilistic divergence between predicted and ground truth token (word index in codebook) sequences:
\begin{equation}
L_{\text{text}} = -\sum_{i=1}^{N} y_{\text{true},i} \log(y_{\text{pred},i})
\end{equation}
where $N$ denotes sequence length, $y_{\text{true},i}$ represents ground truth tokens, and $y_{\text{pred},i}$ represents predicted token probabilities.

\paragraph{Word Matching Loss.}
To further improve vocabulary precision and text coherence, we introduce a word matching loss $\mathcal{L}_{\text{word}}$ that directly compares predicted and ground-truth token embeddings using cosine similarity. This loss is defined as:
\begin{equation}
\mathcal{L}_{\text{word}} = \frac{1}{B} \sum_{b=1}^{B} (1 - \text{sim}(e_{\text{pred},b}, e_{\text{true},b})),
\end{equation}
where $B$ is the batch size, $e_{\text{pred},b}$ and $e_{\text{true},b}$ represent the embeddings of the predicted and ground-truth tokens for batch sample $b$, respectively, and $\text{sim}(\cdot, \cdot)$ denotes the cosine similarity function. By minimizing this loss, the model learns to align the predicted token embeddings with their true counterparts, enhancing the semantic accuracy and fluency of the generated text. This approach proves particularly effective in addressing over-long or incoherent outputs observed with the text generation loss alone.

\paragraph{Contrastive Learning Loss.}
Inspired by recent multimodal representation learning \cite{chen2020simple,Lu-2023-arXiv-multimodal,Zhang-arxiv-2022-Multimodal}, we introduce a contrastive alignment loss that encourages semantic proximity between pose and text embeddings:
\begin{equation}
L_{\text{contrast}} = -\log \frac{\exp(sim(z_{\text{pose}}, e_{\text{text}}) / \tau)}{\sum_{j} \exp(sim(z_{\text{pose}}, e_{\text{text}}^j) / \tau)}.
\end{equation}
The loss leverages cosine similarity between pose embeddings $z_{\text{pose}}$ (output of module) and text embeddings $e_{\text{text}}$  (index of codebook), with $\tau$ as a temperature scaling parameter to control embedding separation, where $\exp(\cdot)$ denotes the exponential function.

  \paragraph{Composite Loss Formulation.}
  The total loss is formulated as a weighted combination of these components:
  \begin{equation}
  \begin{split}
  L_{\text{total}} = &\lambda_{\text{text}} L_{\text{text}} \\
  &+ \lambda_{\text{word}} L_{\text{word}} + \lambda_{\text{contrast}}
  L_{\text{contrast}}.
  \end{split}
  \end{equation}

\begin{wraptable}{r}{0.48\textwidth}
  \centering
  \vspace{-22pt}
  \resizebox{\linewidth}{!}{
  \begin{tabular}{lcc}
  \toprule
  \text{Config} & \text{Stage 1 (Pose2Gloss)} & \text{Stage 2 (Video2Pose)} \\
  \midrule
  \text{optimizer} & \text{AdamW} & \text{AdamW} \\
  \text{base learning rate} & $1 \times 10^{-3}$ & $1 \times 10^{-3}$ \\
  \text{weight decay} & 0.01 & 0.01 \\
  \text{optimizer momentum} & $\beta_1 = 0.9$, $\beta_2 = 0.999$ & $\beta_1 = 0.9$, $\beta_2 = 0.999$ \\
  \text{learning rate schedule} & \text{Cosine decay (min = $0.05 \times lr$)} & \text{Cosine decay (min = $0.01 \times lr$)} \\
  \text{warmup steps} & \text{Dynamic (2\%-10\% of total steps)} & 10\% of total steps \\
  \text{training epochs} & 500--1000 & 300 \\
  \text{batch size} & 32 & 32 \\
  \text{gradient accumulation} & 4 & 0 \\
  \text{gradient clipping} & 7.0 & 1.0 \\
  \text{label smoothing} & 0.1 & N/A \\
  \text{teacher forcing ratio} & \text{Decay from 0.5 to 0} & N/A \\
  \text{text loss weight} & 1 & 1 \\
  \text{word match weight} & 1 & N/A \\
  \text{contrast loss weight} & 1 & N/A \\
  \text{max sequence length} & 64 & 64 \\
  \text{temperature} & 0.7 (inference) & 0.7 (inference) \\
  \text{beam size} & 5 (inference) & 5 (inference) \\
  \text{length penalty} & 0.7 (inference) & 0.7 (inference) \\
  \text{top-k} & 50 (sampling) & 50 (sampling) \\
  \text{top-p} & 0.92 (sampling) & N/A \\
  \text{repetition penalty} & 2.5 (beam search) & N/A \\
  \bottomrule
  \end{tabular}
  }
  \caption{\textbf{Comprehensive training configurations and hyperparameters:} Optimizer settings, learning schedules, training parameters, and inference configurations for both stages.}
  \label{tab:training_recipe_updated}
  \vspace{-20pt}
\end{wraptable}

\subsection{Stage 2: ViT Video2Pose Part Training}

Following feature transfer learning principles \cite{yosinski2014transferable}, we freeze the Pose2Gloss module and train the Video2Pose component by minimizing the reconstruction error:
\begin{equation}
L_{\text{video2pose}} = \sum_{i} ||\hat{P}_i - P_i||^2,
\end{equation}
where $\hat{P}_i$ represents predicted pose representations and $P_i$ denotes ground truth pose representations.
This 2-stage approach enables specialized learning for each module, creating a robust sign recognition system that effectively bridges visual and textual domains, ultimately achieving a synergistic effect where the combined performance exceeds the sum of its parts.

\subsection{Implementation Details of Latent Space}

For Stages 1 and 2, we use the linguistically annotated ASLLRP SignStream® 3 Corpus of continuous signing \cite{neidle2022asl} as primary datasets for American Sign Language (ASL) recognition. 
During Stage 1, we initially found that the heterogeneity and varying formats of pose features from diverse sources had a significant negative impact on the structural integrity of our latent space. To address this, we prioritized optimizing the cross-modal alignment and feature consistency in the Pose2Gloss module (Stage 1) through a gloss-aware latent distillation loss.
However, we observed that relying solely on feature-level distillation did not sufficiently address the nuances of sign-to-text generation quality.

We explored text generation using a cross-entropy loss $\mathcal{L}_{\text{text}}$
. While this approach improved basic text output, it frequently generated over-long and incoherent text, limiting practical utility. To address this, we investigated an n-gram consistency loss designed to balance precision and recall, but found this implicit matching approach less effective than directly focusing on vocabulary alignment. We therefore introduced a word matching loss that measures cosine similarity between predicted and ground-truth token embeddings, encouraging precise vocabulary generation within our custom CodeBook. Additionally, we incorporated contrastive learning with temperature parameter $\tau$ to further encourage semantic proximity between the ViT-constructed pose latent space and text representations.

For Stage 2 (Video2Pose), we froze the Pose2Gloss parameters trained in Stage 1 and developed a Vision Transformer-based model to convert raw video inputs directly into the unified pose latent representation, minimizing the reconstruction loss $\mathcal{L}_{\text{video2pose}} = \sum_{i} ||\hat{P}_i - P_i||^2$, where $\hat{P}_i$ represents predicted latent pose representations and $P_i$ denotes ground-truth unified poses. This 2-stage approach, leveraging the pre-trained ViT latent space, ensures robust video-to-pose conversion while maintaining evaluation efficiency.
These optimizations, combined with gradient accumulation,
as detailed in Table~\ref{tab:training_recipe_updated}, resolved initial training instability and enhanced model performance. The best options for weights, pose types, and temperature values for different losses were determined through extensive ablation experiments.

\section{More Continuous Sign Language Recognition Details}

\paragraph{Hybrid Temporal Modeling.} The distilled latent sequences feed a hybrid backbone that remains modality-agnostic. First, a \textit{multi-scale TemporalConv stack} captures hierarchical temporal patterns through parallel 1D convolutions with kernel sizes $\mathcal{K} = \{3, 5, 7\}$, corresponding to short-term (3 frames $\approx$ 0.1s), medium-term (5 frames $\approx$ 0.17s), and long-term (7 frames $\approx$ 0.23s) dependencies. Each branch applies:
\begin{equation}
\begin{aligned}
h_t^{(k)} = \text{MaxPool}(\text{ReLU}( & \text{BN}(\text{Conv1D}_k(z_t)))), \\
& k \in \mathcal{K}
\end{aligned}
\end{equation}
where stride-2 pooling progressively compresses the sequence from $T$ to $T' \approx T/4$. The outputs are concatenated and projected to $h_t \in \mathbb{R}^{1024}$. These features then pass through a bidirectional LSTM (2 layers, hidden size 512 per direction) to capture long-range dependencies, yielding $u_t \in \mathbb{R}^{1024}$. This design mirrors the receptive field hierarchy in human visual processing, where ASL comprehension relies on both rapid hand movements and sustained body postures.

\section{More Experimental Results}

\subsection{Evaluation Metrics}
\label{subsec:more_metric}

For sign recognition, we measure: 

\noindent\textbf{1. Word Error Rate (WER):} 
As the standard metric for Continuous Sign Language Recognition (CSLR), WER quantifies the alignment between the predicted gloss sequence and the ground truth by calculating the normalized edit distance:
\begin{equation}
    \text{WER} = \frac{S + D + I}{N}
\end{equation}
where $S$, $D$, and $I$ represent the number of \textit{substitutions}, \textit{deletions}, and \textit{insertions} required to transform the hypothesis into the reference, respectively. $N$ denotes the total number of glosses in the reference. A lower WER indicates higher recognition stability, particularly in capturing the subtle transitions between sign gestures.

\noindent\textbf{2. BLEU-n Score:}
To evaluate the translation quality of the generated glosses or sentences, we employ the BLEU score, which measures the $n$-gram precision between the candidate and reference sequences:
\begin{equation}
    \text{BLEU} = \text{BP} \cdot \exp \left( \sum_{n=1}^{k} w_n \ln p_n \right)
\end{equation}
Here, $p_n$ denotes the modified $n$-gram precision, and $w_n$ represents the uniform weight $1/k$. The Brevity Penalty ($\text{BP}$) is incorporated to penalize overly short predictions:
\begin{equation}
    \text{BP} = \begin{cases} 1 & \text{if } c > r \\ e^{(1-r/c)} & \text{if } c \leq r \end{cases}
\end{equation}
where $c$ is the length of the candidate and $r$ is the reference length. Higher BLEU scores reflect better semantic consistency and linguistic fluency of the SignX decoder.

\noindent\textbf{3. Per-Instance (P-I) Top-1 Accuracy:} 
Per-instance (P-I) accuracy is the primary metric used to evaluate the overall recognition performance of \textbf{SignX} across the entire test set. It treats each video sample as an independent and equal evaluation unit, representing the global classification accuracy (Micro-average). Formally, it is defined as:
\begin{equation}
    \text{Acc}_{P\text{-}I} = \frac{\sum_{i=1}^{C} n_{i,i}}{\sum_{i=1}^{C} N_i}
\end{equation}
where $C$ denotes the total number of sign categories in the dataset (e.g., 2,000 for WLASL2000), $n_{i,i}$ is the number of samples from class $i$ correctly identified by the model, and $N_i$ is the total number of test samples belonging to category $i$. 

\begin{figure}[t]
\vspace{-4pt}
    \centering
    \includegraphics[width=0.99\linewidth]{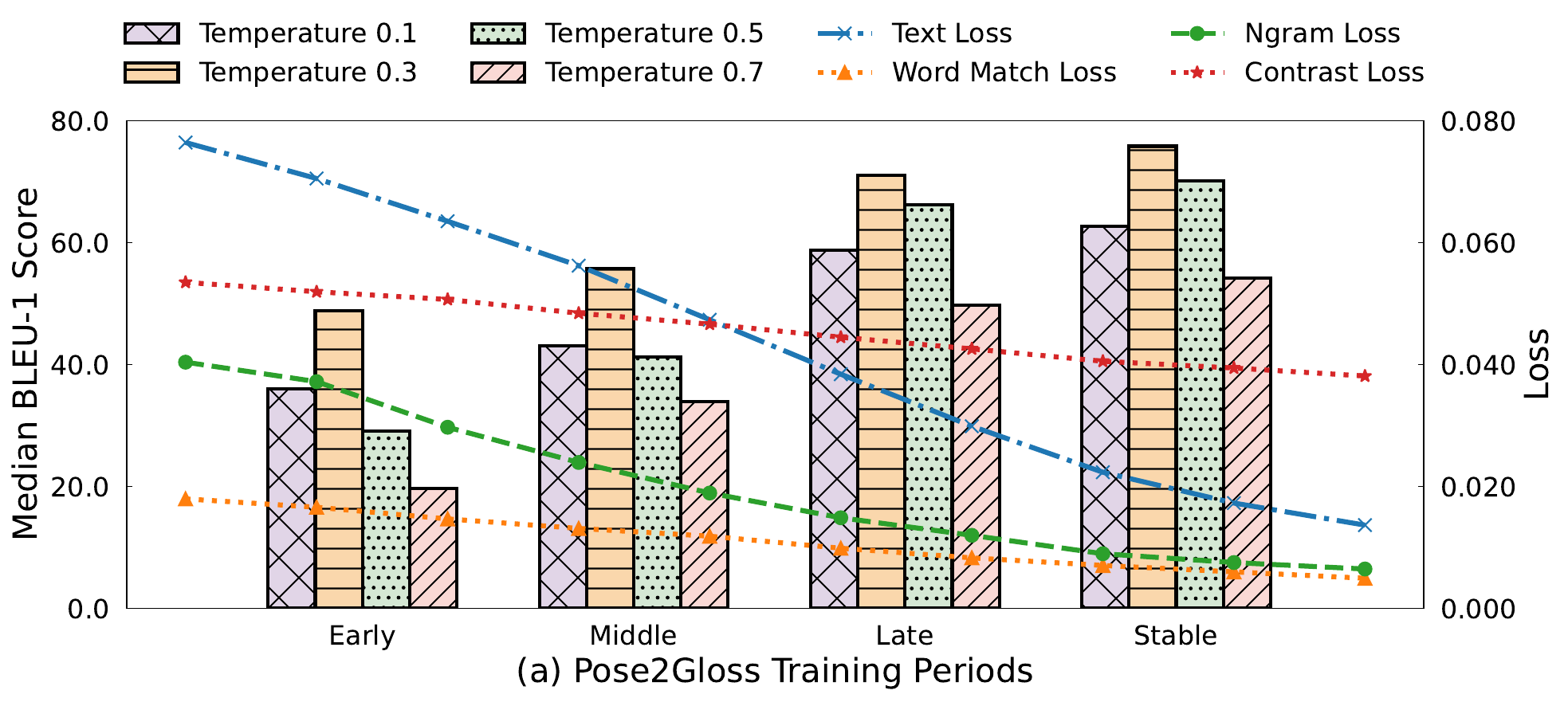}
    \vspace{-4pt}
    \caption{\textbf{Optimization efficiency in constructing the pose-rich latent space.} This figure tracks the convergence of four critical loss components designed to shape the latent space. The synchronized decline of Text and Word Match losses demonstrates the successful encoding of sign semantics into the unified latent representation.}
    \label{fig:latent_space_convergence}
\vspace{-4pt} 
\end{figure}

\begin{table}[t]

  \scriptsize
  \centering

  \resizebox{0.88\textwidth}{!}{
  \begin{tabular}{>{\raggedright\arraybackslash}p{\linewidth}}

  \toprule

  \textbf{Phoenix-14T (German Sign Language)} \\

  \vspace{2pt}

  \textit{Generated Text:} ES-BEDEUTET HERBST WOLKE UND UND KOENNEN GEWITTER \\

  \textit{English:} It means autumn cloud and and can thunderstorm \\

  \textit{Ground Truth:} ES-BEDEUTET VIEL WOLKE UND KOENNEN REGEN GEWITTER KOENNEN \\

  \textit{English:} That means lots of clouds and again and again strong showers and thunderstorms \\

  \cmidrule{1-1}

  \textit{Generated Text:} TAG DANN SUED SUEDOST NACH MITTAG IX NOVEMBER GEWITTER \\

  \textit{English:} Day then south southeast after midday IX November thunderstorm \\

  \textit{Ground Truth:} TAG DANN SUED SUEDOST REGEN NACH MITTAG IX AUCH GEWITTER \\

  \textit{English:} During the day it rains in the south and south-east, and in the afternoon there are also local thunderstorms \\

  \midrule

  \textbf{CSL-Daily (Chinese Sign Language)} \\

  \vspace{2pt}

  \textit{Generated Text:} 这\ 这\ 考虑\ 画画\ 是\ 自己\ 仔细\ 方法\ 分析 \\

  \textit{English:} This, this, consider painting is one's own careful method analysis \\

  \textit{Ground Truth:} 这\ 画画\ 是\ 他\ 自己\ 努力\ 仔细\ 想\ 方法\ 结果 \\

  \textit{English:} This painting is the result of his own hard work and careful thinking \\

  \cmidrule{1-1}

  \textit{Generated Text:} 不管\ 结果\ 情况\ 情况\ 大概\ 努力 \\

  \textit{English:} No matter the result, the situation, the situation, or the effort \\

  \textit{Ground Truth:} 不管\ 结果\ 情况\ 我\ 一直\ 努力 \\

  \textit{English:} No matter the result or situation, I will always work hard \\

  \bottomrule

  \end{tabular}
  }

  \vspace{8pt}

  \caption{\textbf{Qualitative evaluation of sign language translation.} 
  The examples demonstrate the model's generalization capability across diverse sign languages (German and Chinese).}

  \label{tab:qualitative}

  \vspace{-10pt}

\end{table}

\subsection{Training Efficiency of Constructing Compact Pose-Rich Latent Space}
\label{subsec:Training_Efficiency_of_Latent_Space}
The efficiency of constructing our pose-rich latent space is quantitatively analyzed in Fig.~\ref{fig:latent_space_convergence}, which monitors the progression of four specialized loss objectives: Text Loss, Word Match Loss, Ngram Loss, and Contrast Loss. The line plots illustrate a consistent downward trend in Text and Word Match losses, confirming that the latent space is effectively learning to accommodate high-dimensional pose information while maintaining precise semantic mappings. The stabilization of Ngram Loss and the controlled trajectory of Contrast Loss further indicate that the space is achieving structural coherence, ensuring that heterogeneous pose features are harmonized into a compact, information-dense representation.

\subsection{Qualitative Assessment}
\label{subsec:qualitative}

As illustrated in Table \ref{tab:qualitative}, the model demonstrates robust performance across both weather broadcast (Phoenix-14T) and daily conversation (CSL-Daily) domains. In the Phoenix-14T examples, the model successfully captures the core semantic content of weather forecasts. In the first example, key meteorological terms like ``WOLKE'' (cloud) and ``GEWITTER'' (thunderstorm) are accurately predicted, though with minor lexical substitutions (e.g., ``HERBST'' for ``VIEL''). The second example shows the model's ability to preserve temporal markers (``NACH MITTAG'') and directional information (``SUED SUEDOST''), while occasionally omitting modifiers like ``AUCH'' (also) and ``REGEN'' (rain). For CSL-Daily, the model exhibits strong semantic preservation despite surface-level variations. The first example accurately conveys the relationship between artistic creation and effort, with key concepts like ``画画'' (painting), ``仔细'' (careful), and ``方法'' (method) correctly identified. The second example demonstrates the model's capability in handling abstract concepts, where ``不管'' (no matter) and ``努力'' (effort) are preserved, though with slight word duplication (``情况\ 情况'') that does not significantly impair comprehension. These results indicate that our approach effectively maintains semantic coherence across diverse sign language contexts, though there remains room for improvement in handling function words and fine-grained temporal/spatial modifiers.

\begin{figure*}[t]
\vspace{-4pt}
    \centering
    \includegraphics[width=0.99\linewidth]{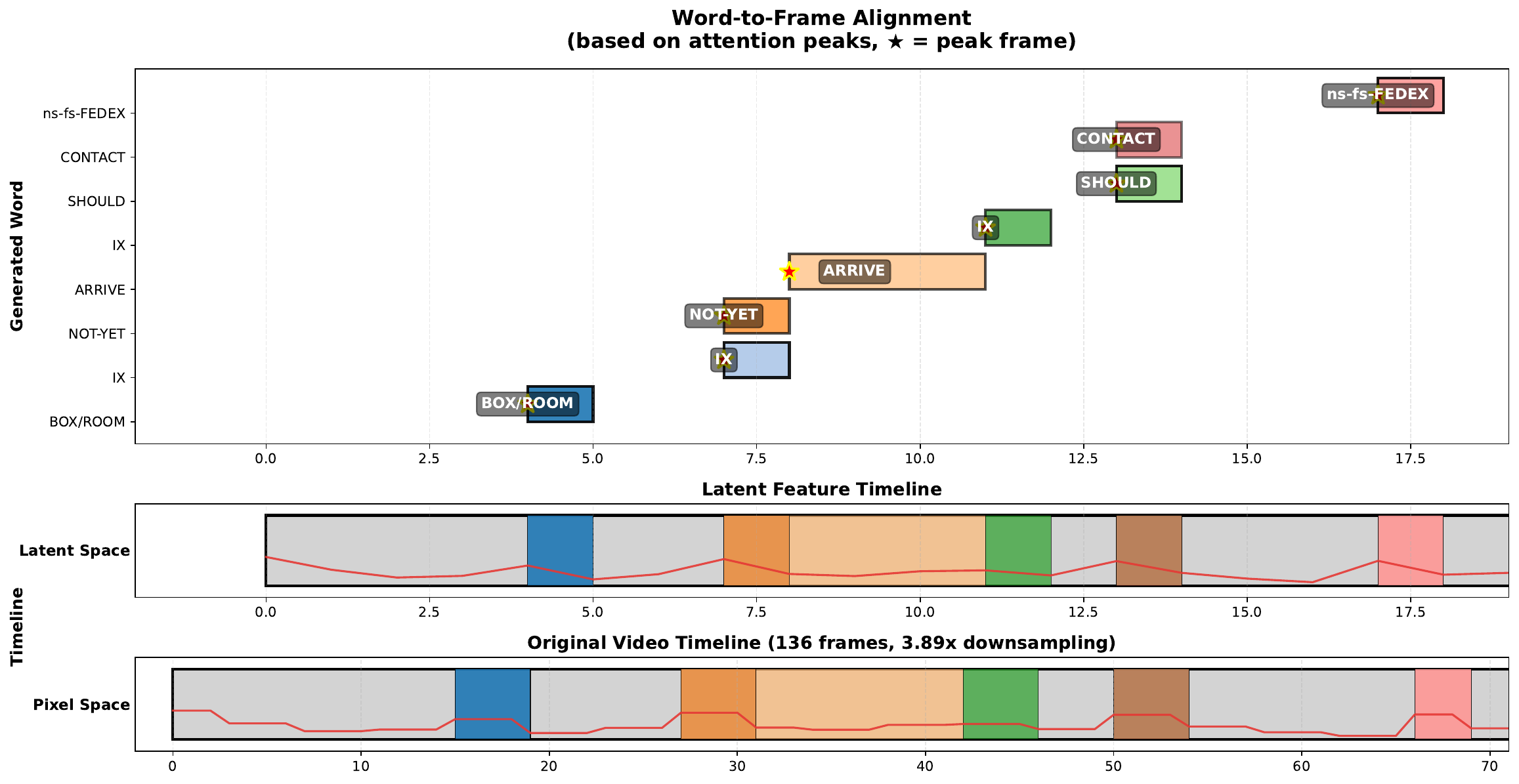}
    \vspace{-12pt}
    \caption{\textbf{Attention Analysis.} Here, we have analyzed the feature extraction and recognition section. The above figure represents the analyzed Gloss for different potential representations. Each representation is derived from 5 to 10 frames. This area represents the relative position of this representation. The two progress bars below represent the relative positions at the feature level and the pixel frame level respectively. The red line is the intensity of attention.}
    \label{fig:attention1}
\vspace{-4pt} 
\end{figure*}

\begin{figure*}[t]
\vspace{-4pt}
    \centering
    \includegraphics[width=0.99\linewidth]{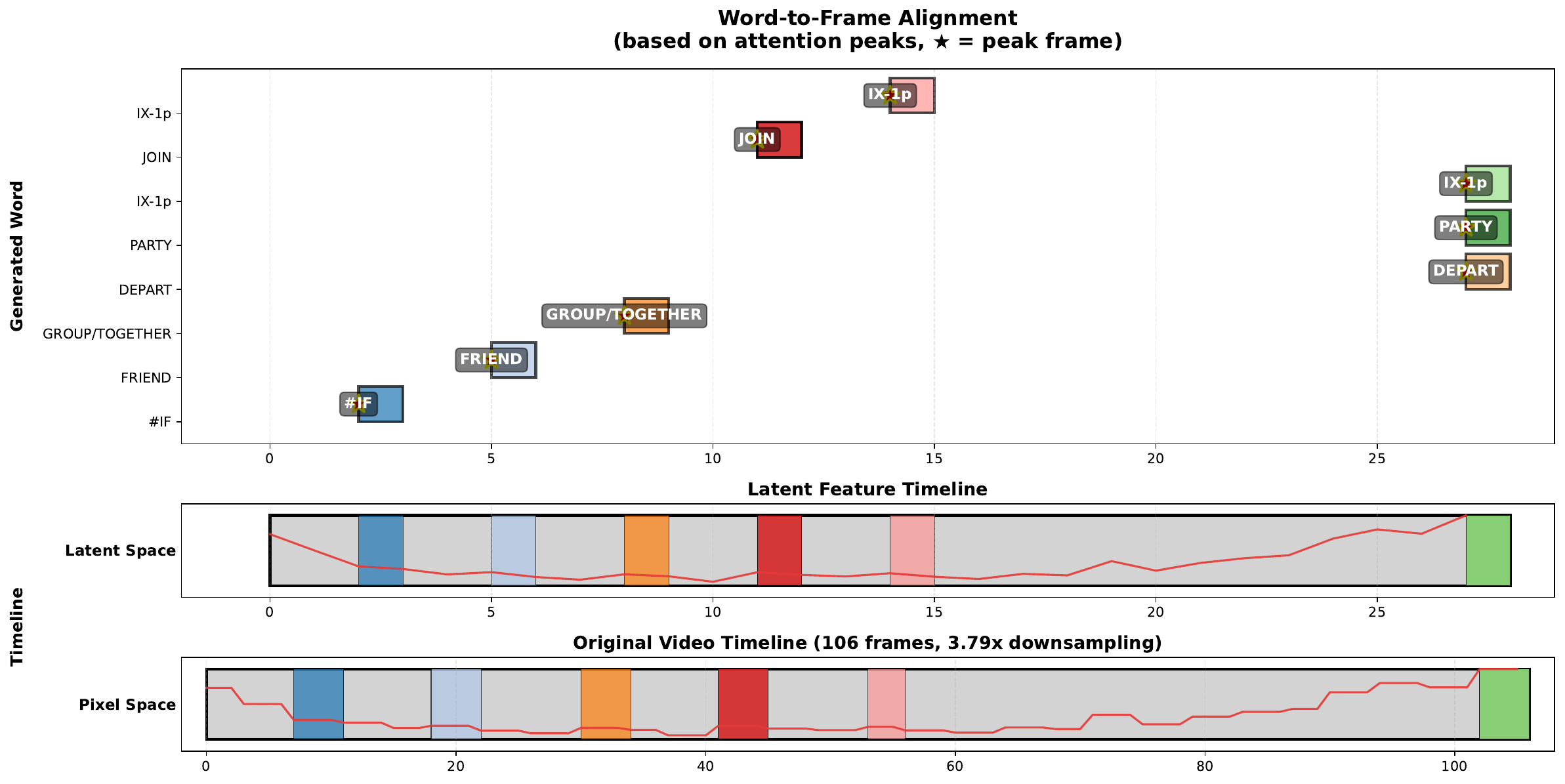}
    \vspace{-12pt}
    \caption{\textbf{Further analysis of the reasons for excellence.} To verify the entire pipeline of our model for the CSLR process and understand how its mechanism operates, we selected a sample in the translation process where all video features were concentrated at the relatively tail end. In this sample, the last video feature encoded information about three postures.}
    \label{fig:attention2}
\vspace{-4pt} 
\end{figure*}

\subsection{Visualization and in-depth Analysis (I)}

To analyze the internal mechanics of our model, we visualize the temporal alignment in Figure~\ref{fig:attention1}. In our framework, one latent feature typically contains information from approximately 5 to 10 video frames. Consequently, a single gloss is often represented by only one or two features.

As shown in the three-layer alignment (Gloss $\leftrightarrow$ Feature Index $\leftrightarrow$ Original Frames), we observe a key phenomenon: even a single feature can be identified with multiple meanings. For example, \textbf{Peak Feature 6} is simultaneously mapped to the glosses \textit{IX} and \textit{NOT-YET}. This proves that even when the data is compressed into a single latent variable, it still retains highly useful and multifaceted information. This high information density allows the model to maintain accurate recognition while significantly reducing the computational burden.

Furthermore, for this overlapping situation, we have found that our Latent CSLR model can actually restore the chronological order of them. There are two possible scenarios. One is that even in the Latent space, the information represented has a hidden sequence. The other is that the Latent CSLR model restored this sequence relationship. Regardless of which situation it is, it demonstrates the excellence of our model. We can further analyze in Section \ref{subsec:vis2} to determine which scenario it is.

\subsection{Visualization and in-depth Analysis (II)}
\label{subsec:vis2}


In the previous section, we discovered that our Latent Space (Vid2Pose) possesses excellent compression capabilities, where a single feature can represent multiple pose details. However, it remained unclear whether this success was because temporal information was recorded during compression or if the Latent CSLR restored the temporal information during translation.

Therefore, in Figure \ref{fig:attention2}, we provide a sample where the features are primarily distributed at the end of the video. In this case, three glosses are output by the model only after it has finished viewing the video. We surprisingly found that this represents two major advantages of our model:
\begin{enumerate}
    \item \textbf{Thinking-based Compression:} When Video2Pose provides features, it confirms and outputs them only after seeing the full video if it cannot determine them immediately. \textbf{Furthermore}, as the video nears its end, the level of attention begins to increase, which is also evidence that the model has thought through and provided the video's features after careful consideration.
    \item \textbf{Contextual Restoration:} Once our Latent CSLR obtains these features, it restores them to their correct temporal positions. \textbf{For example}, the final output of the model is ``\#IF FRIEND GROUP/TOGETHER DEPART PARTY IX-1p JOIN IX-1p'', while the correct answer is ``\#IF FRIEND GROUP/TOGETHER GO-OUT PARTY IX-1p JOIN IX-1p''. We can observe that after careful consideration, the feature "PARTY", which was initially determined later, was successfully returned to its correct position. This is a remarkable ability. Although there was a spelling mistake in one word, it was not caused by our innovative approach, but rather due to the insufficient capabilities of the existing ViT.
\end{enumerate}


This suggests that the model effectively internalizes a temporal realignment mechanism during the optimization process, enabling it to compensate for feature manifestation latency and accurately restore the objective chronological sequence. Based on these two experiments, we now know \textbf{why our model is excellent} (intelligent, ``thinking'' compression) and \textbf{where it excels} (the Latent CSLR translates by combining context). \textbf{We have discovered the reason for this success}—learning from feature post-positioning during training—and validated the future prospects and superiority of our model.

\section{Discussion}

Here we discuss some common issues:

\subsection{Integrity of Technical Implementation}
The technical details described in Section \ref{sec:methodology}, such as adaptive feature pruning and latent space organization, serve as the foundational implementation of our Latent CSLR framework. These components are not disjointed tricks; rather, they have been validated as a holistic system in our ablation studies (As shown in Fig. \ref{fig:ablation_study}, \ref{fig:latent_space_convergence}, \ref{fig:latent_representation_quality}, \ref{fig:pose_dimension_heatmap} and Table. \ref{tab:modality_ablation_final}) to ensure the structural integrity of the latent space. 

\subsection{Rigorous Benchmarking on the ASLLRP Dataset}
As ASLLRP is a relatively new dataset for the CSLR task, there is a lack of publicly available baseline results from modern models. To ensure a fair comparison under identical pose input conditions, we re-implemented and fine-tuned several representative state-of-the-art (SOTA) frameworks, including SLTUNET \cite{zhang2023sltunet} and C2SLR \cite{zuo2022c2slr}. Our selection criteria prioritized mainstream methods that represent established technical trajectories in sign language understanding, rather than solely focusing on the most recent publication dates.

\subsection{Component Customization and Scalability}
All architectural components in SignX have been deeply modified to serve the objective of latent space recognition. For instance, our Custom CodeBook is utilized to learn and initialize pose-rich features beneficial for linguistic mapping. This design is inherently scalable to an unlimited vocabulary and does not restrict the model's capacity for future expansion. Furthermore, while our framework utilizes a comprehensive set of modalities, these modules are highly reusable across different datasets. Compared to ultra-large-scale models such as Uni-Sign \cite{li2025uni}, our approach achieves superior performance with a smaller data footprint, demonstrating a data-frugal and efficient research path.

\subsection{Compactness and Computational Efficiency}
The compactness of our latent space has been strictly validated through ablation evaluations (As shown in Fig. \ref{fig:ablation_study}, \ref{fig:latent_space_convergence}, \ref{fig:latent_representation_quality}, \ref{fig:pose_dimension_heatmap} and Table. \ref{tab:modality_ablation_final}). Our compression is measured against the original raw video frame dimensionality of 150,528 ($224 \times 224 \times 3$). Through adaptive feature pruning, we successfully stabilize the latent width at approximately 1024 (The initial dimension after padding is 2048, and after pruning, compression and organization, it becomes 1024.), achieving a compression ratio exceeding 140x\footnote{Considering that multiple frames are compressed into one feature, our effective compression rate will be even higher. More importantly, after we have completed the training, we no longer need the posture extraction data preprocessing tool. In the long run, this saves a lot of time for learning posture information.}. This significant reduction effectively mitigates computational overhead while maintaining high information density, as substantiated by our inference efficiency analysis showing a 50-fold acceleration compared to traditional pixel-space baselines (As shown in Table \ref{tab:efficiency}).

\subsection{Discussion on the overall motivation and methodology}

Overall, our key insight is that not only should we perform generation tasks in the Latent Space, but we should also perform recognition tasks in the Latent Space. This was deemed successful by the experiments. Initially, we had the option of using pre-trained components such as VAE \cite{kingma2013auto,dong2022act} and DINO3 \cite{simeoni2025dinov3} as alternatives, but they lacked pose-level information, lacked effective sign language semantic information, and the simple CNN in VAE could not meet our needs for scalability and stability. Therefore, after careful consideration, we adopted the existing methods. In the future, we will make more resources available, such as code and preprocessed data, to support the sign language research community.

\end{CJK*}
\end{document}